\documentclass[runningheads]{llncs}
\usepackage[T1]{fontenc}
\usepackage{graphicx} 
\usepackage{xcolor}
\usepackage{listings}
\usepackage{hyperref}
\usepackage{enumitem}
\usepackage{booktabs}
\usepackage{multicol}
\usepackage{xurl}
\usepackage[htt]{hyphenat}
\usepackage{breakcites}

\definecolor{codegreen}{rgb}{0,0.6,0}
\definecolor{codegray}{rgb}{0.5,0.5,0.5}
\definecolor{codepurple}{rgb}{0.58,0,0.82}
\definecolor{backcolour}{rgb}{0.95,0.95,0.92}

\newcommand{\KERN}{{\sc**kern} }

\def \eg {\emph{e.g.}, }
\def \etal {\emph{et al.}}

\lstdefinelanguage{json}{
    basicstyle=\normalfont\ttfamily,
    numbers=left,
    numberstyle=\scriptsize,
    stepnumber=1,
    numbersep=8pt,
    showstringspaces=false,
    breaklines=true,
    frame=lines,
    backgroundcolor=\color{backcolour},
    literate=
     *{0}{{{\color{black}0}}}{1}
      {1}{{{\color{black}1}}}{1}
      {2}{{{\color{black}2}}}{1}
      {3}{{{\color{black}3}}}{1}
      {4}{{{\color{black}4}}}{1}
      {5}{{{\color{black}5}}}{1}
      {6}{{{\color{black}6}}}{1}
      {7}{{{\color{black}7}}}{1}
      {8}{{{\color{black}8}}}{1}
      {9}{{{\color{black}9}}}{1}
      {:}{{{\color{codegreen}{:}}}}{1}
      {,}{{{\color{codegreen}{,}}}}{1}
      {\{}{{{\color{black}{\{}}}}{1}
      {\}}{{{\color{black}{\}}}}}{1}
      {[}{{{\color{black}{[}}}}{1}
      {]}{{{\color{black}{]}}}}{1},
}

\lstdefinestyle{mystyle}{
    backgroundcolor=\color{backcolour},
    commentstyle=\color{codegreen},
    keywordstyle=\color{magenta},
    numberstyle=\tiny\color{codegray},
    stringstyle=\color{codepurple},
    basicstyle=\ttfamily\footnotesize,
    breakatwhitespace=false,
    breaklines=true,
    captionpos=b,
    keepspaces=true,
    numbers=left,
    numbersep=5pt,
    showspaces=false,
    showstringspaces=false,
    showtabs=false,
    tabsize=2
}

\begin{document}

\title{A dataset for the recognition of historical and handwritten music scores in western notation}
\titlerunning{A dataset for OMR on historical and handwritten scores in CWMN}
\authorrunning{P. Torras, J. Mayer \etal}

\author{Pau Torras\inst{1,2} \and Ji\v{r}\'{i} Mayer\inst{4} \and Carles Badal\inst{3} \and Martina Dvo\v{r}\'{a}kov\'{a}\inst{5} \and Mark\'{e}ta Herzanov\'{a} Vlkov\'{a}\inst{5} \and Gerard Asbert\inst{1} \and Vojt\v{e}ch Dvo\v{r}\'{a}k\inst{4} \and Samuel \v{S}omorjai\inst{5} \and Jan~Haji\v{c} jr.\inst{4} \and
Alicia Fornés\inst{1,2}}
%
\institute{Computer Vision Center, Barcelona, Spain \\
\email{\{ptorras,afornes\}@cvc.uab.cat}
\and Department of Computer Science, Universitat Autònoma de Barcelona, Spain\\ 
\and Department of Arts and Musicology, Universitat Autònoma de Barcelona, Spain\\
\and Institute of Formal and Applied Linguistics, Charles University, Prague, Czechia  \\ \email{\{mayer,hajicj\}@ufal.mff.cuni.cz}
\and Moravian Library, Brno, Czechia
}

\date{April 2026}

\maketitle

\begin{abstract}
    A large amount of musical heritage has been digitised by memory institutions: libraries, museums, and archives. Nevertheless, the field of Optical Music Recognition (OMR) has struggled with making this music machine-readable, despite advances in deep learning, mostly because no datasets for training systems in realistic conditions were available.
    The MusiCorpus dataset aims to remedy this situation by providing 1,309 pages of historical sheet music, primarily handwritten, with MusicXML transcriptions and symbol annotations. It is the largest dataset of handwritten music to date and the first dataset containing a realistic and representative sample of musical document collections from memory institutions, suitable for training and evaluating both end-to-end and object detection-based OMR systems and comparing their performance.
\end{abstract}

\section{Background \& Summary}

Recovering historical memory enriches society, helping us to understand both our past and our present identity. Knowing the music that we have created over the centuries not only reveals our cultural legacy but also deepens our sense of musical identity. Yet thousands of historical scores—preserved in churches and archives around the world—remain unpublished and unknown. To make this heritage accessible to scholars and the public, it is essential to invest in its recovery through digitization, transcription, analysis, study, dissemination, and recording. Manual transcription, however, is extremely time‑consuming: a single page can require 1.5 to 3 hours of work, depending on its complexity. The sheer volume and diversity of surviving manuscripts make it impossible to transcribe everything by hand, limiting musicology largely to qualitative research. This challenge appears globally and locally. For example, a single major European parish, with over 500 years of musical activity, may hold more than 2,000 works (which means tens of thousands of handwritten pages). In a city like Barcelona, four such chapels exist, with their archives almost entirely preserved. The amount of music awaiting transcription is therefore far beyond what manual efforts can handle, and consequently, the automatic reading of scanned music scores becomes an indispensable solution.

\begin{figure}[ht]
    \centering
    \includegraphics[width=\linewidth]{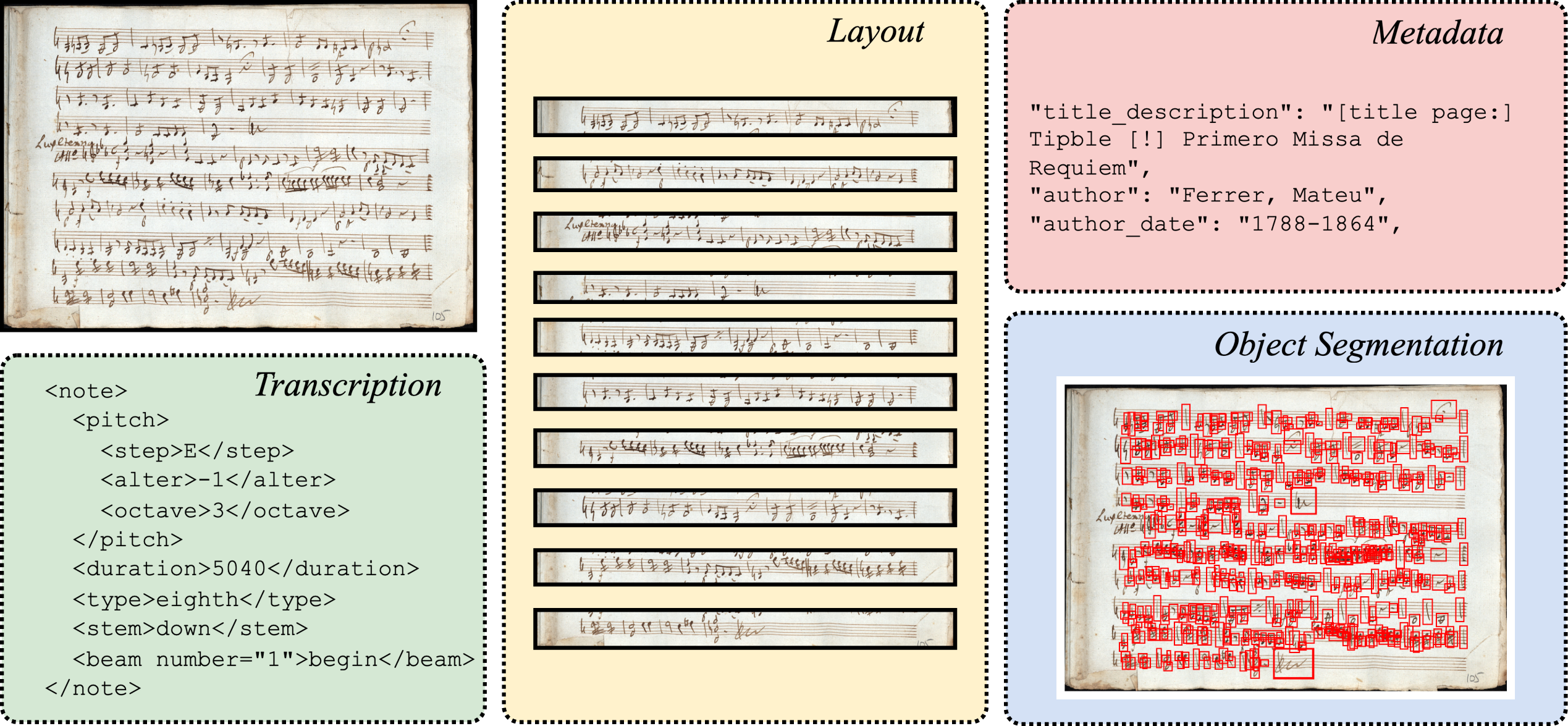}
    \caption{Summary of the MusiCorpus Dataset. It is designed to provide end-to-end transcriptions, layout segmentation, metadata and object-level segmentation records for more than 1.3k pages of music at various levels.}
    \label{fig:abstract}
\end{figure}

\begin{figure}[ht]
    \centering
    \includegraphics[width=0.32\linewidth]{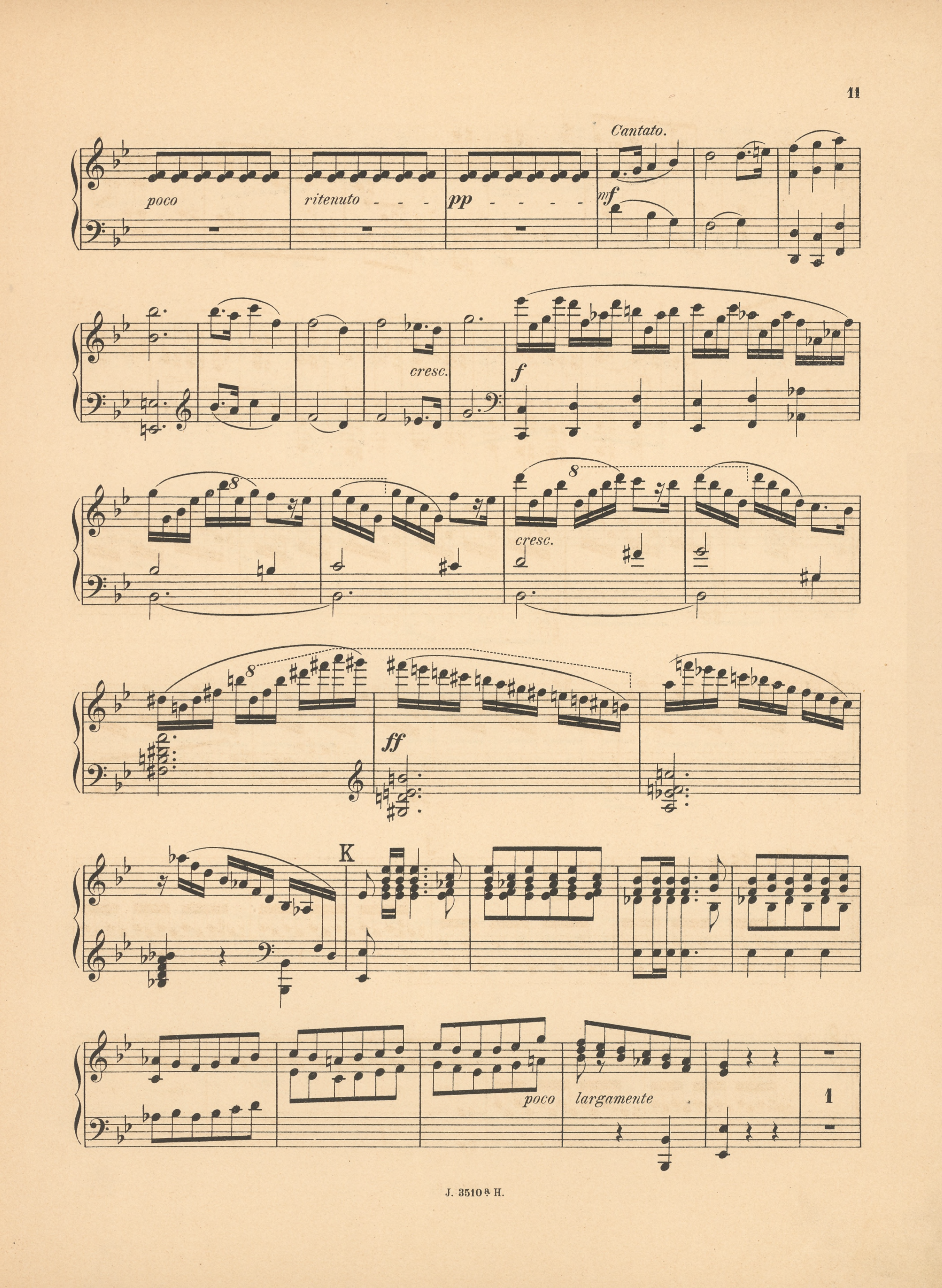}
    \includegraphics[width=0.32\linewidth]{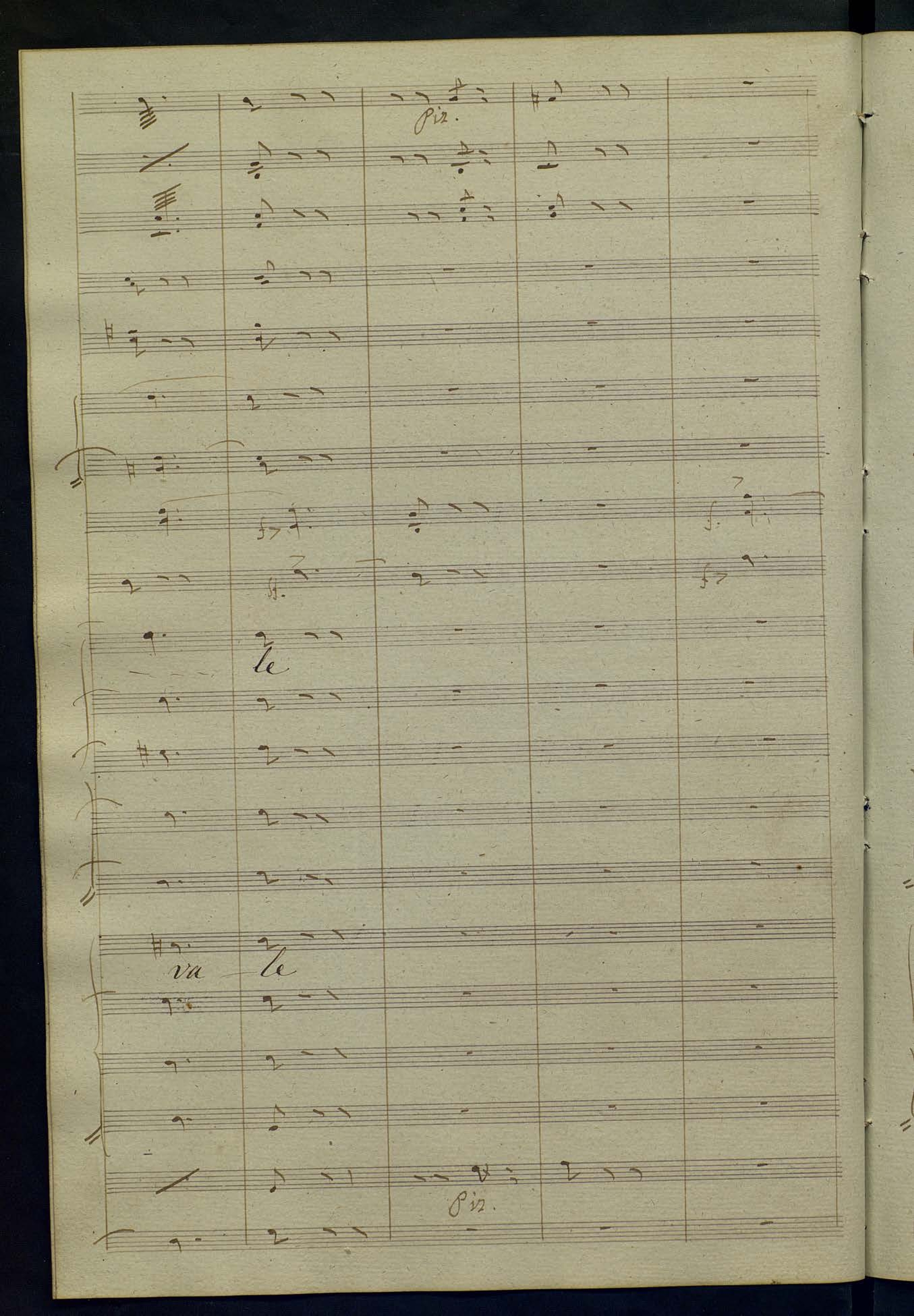}
    \includegraphics[width=0.32\linewidth]{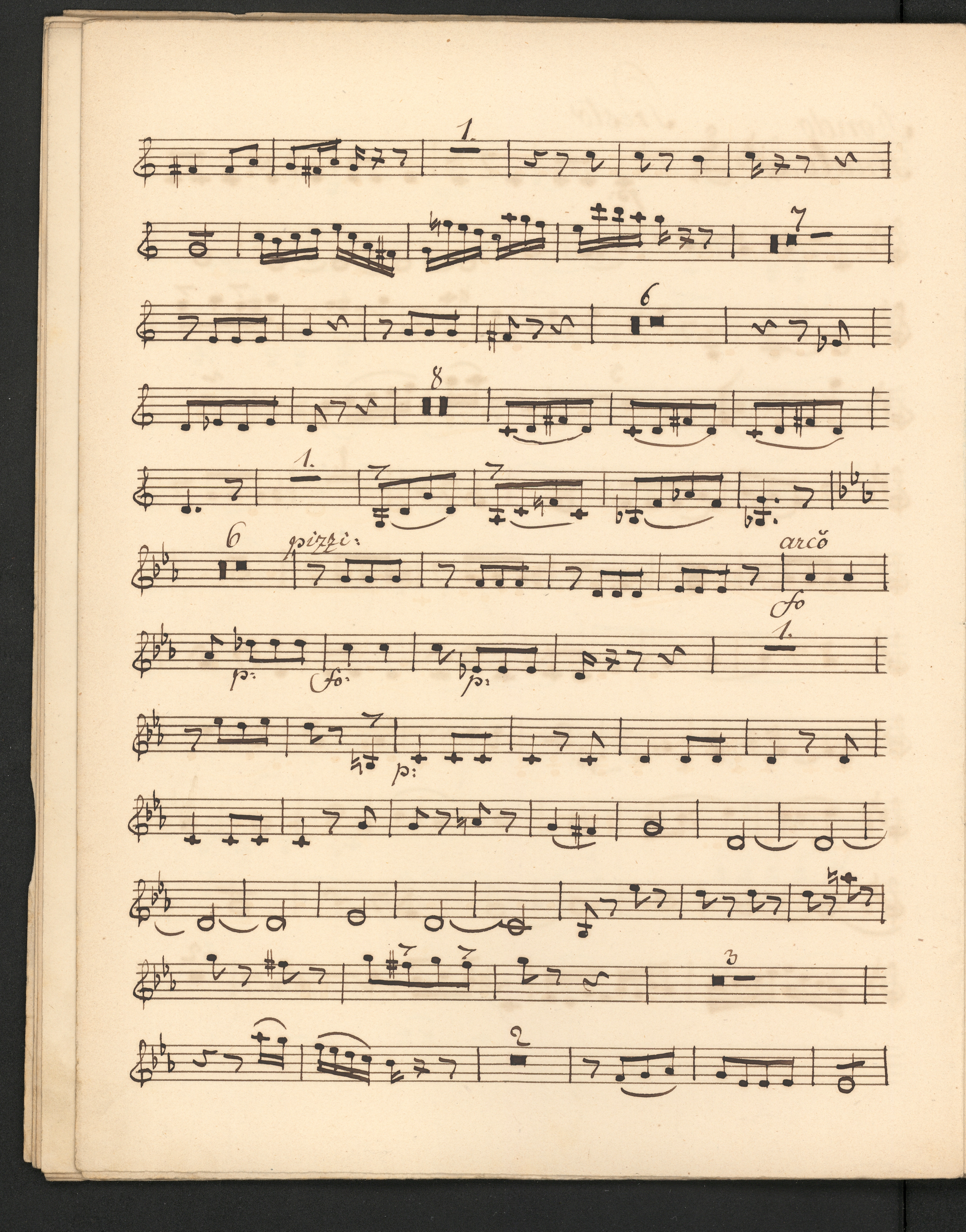}
    \includegraphics[width=0.32\linewidth]{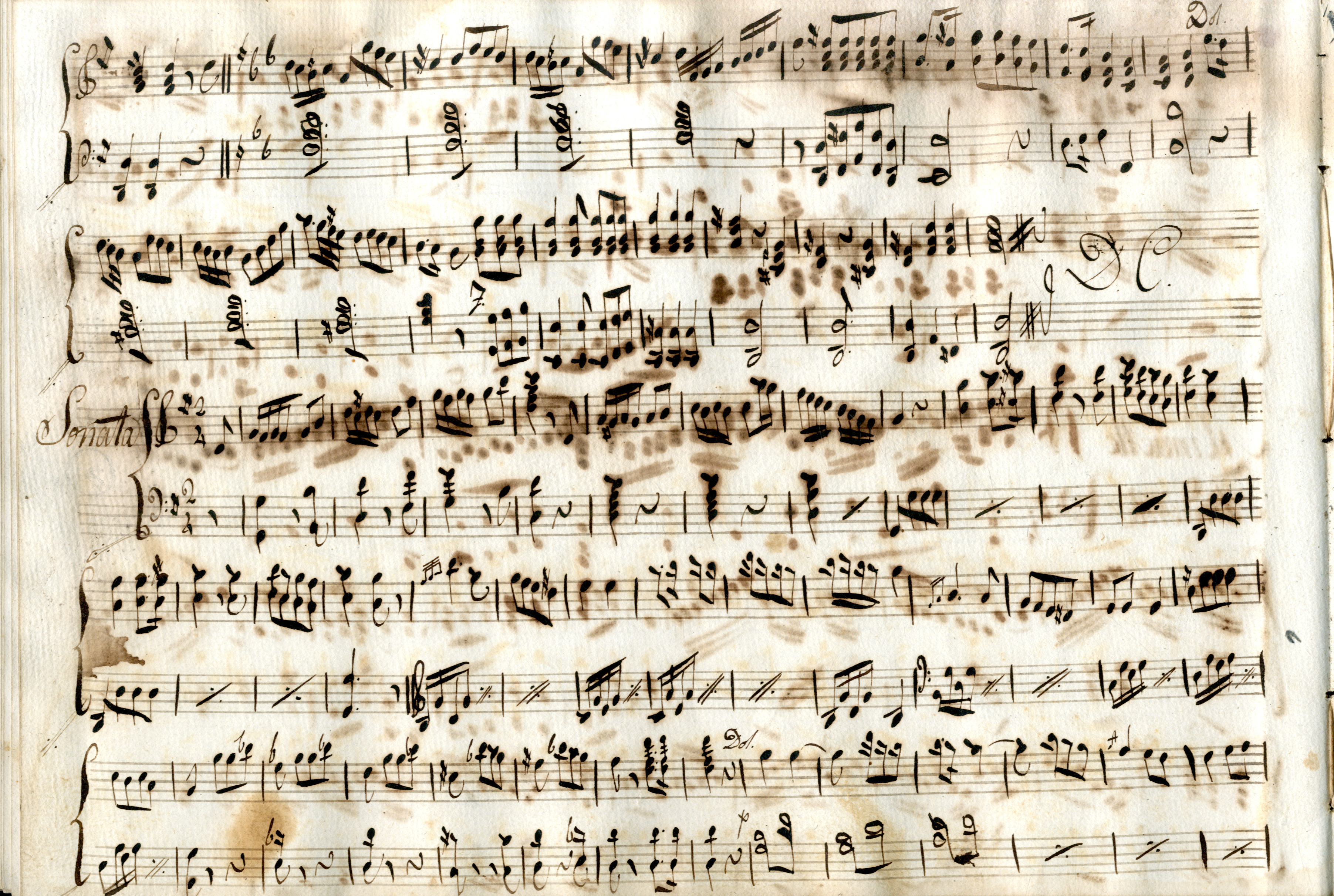}
    \includegraphics[width=0.32\linewidth]{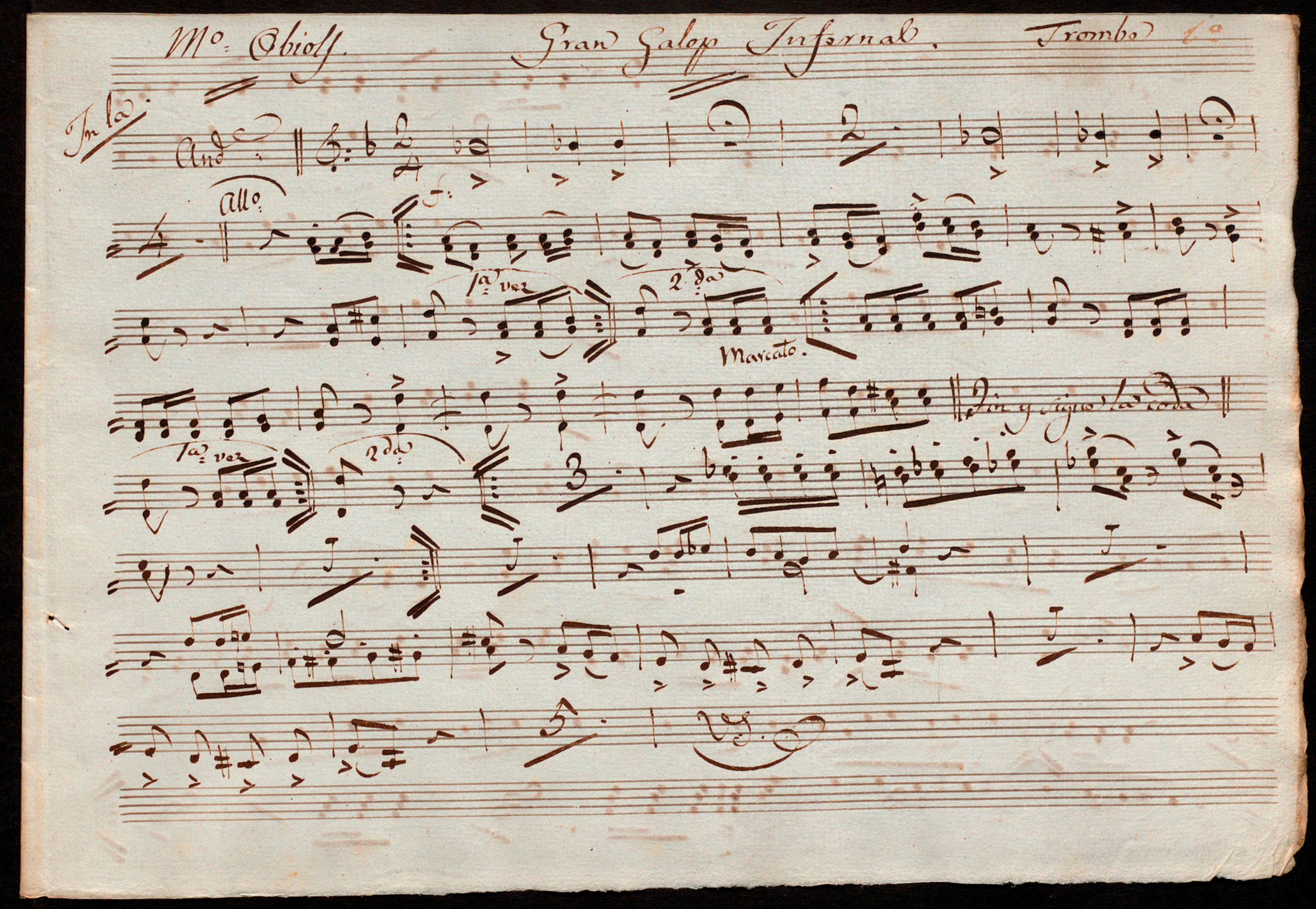}
    \includegraphics[width=0.32\linewidth]{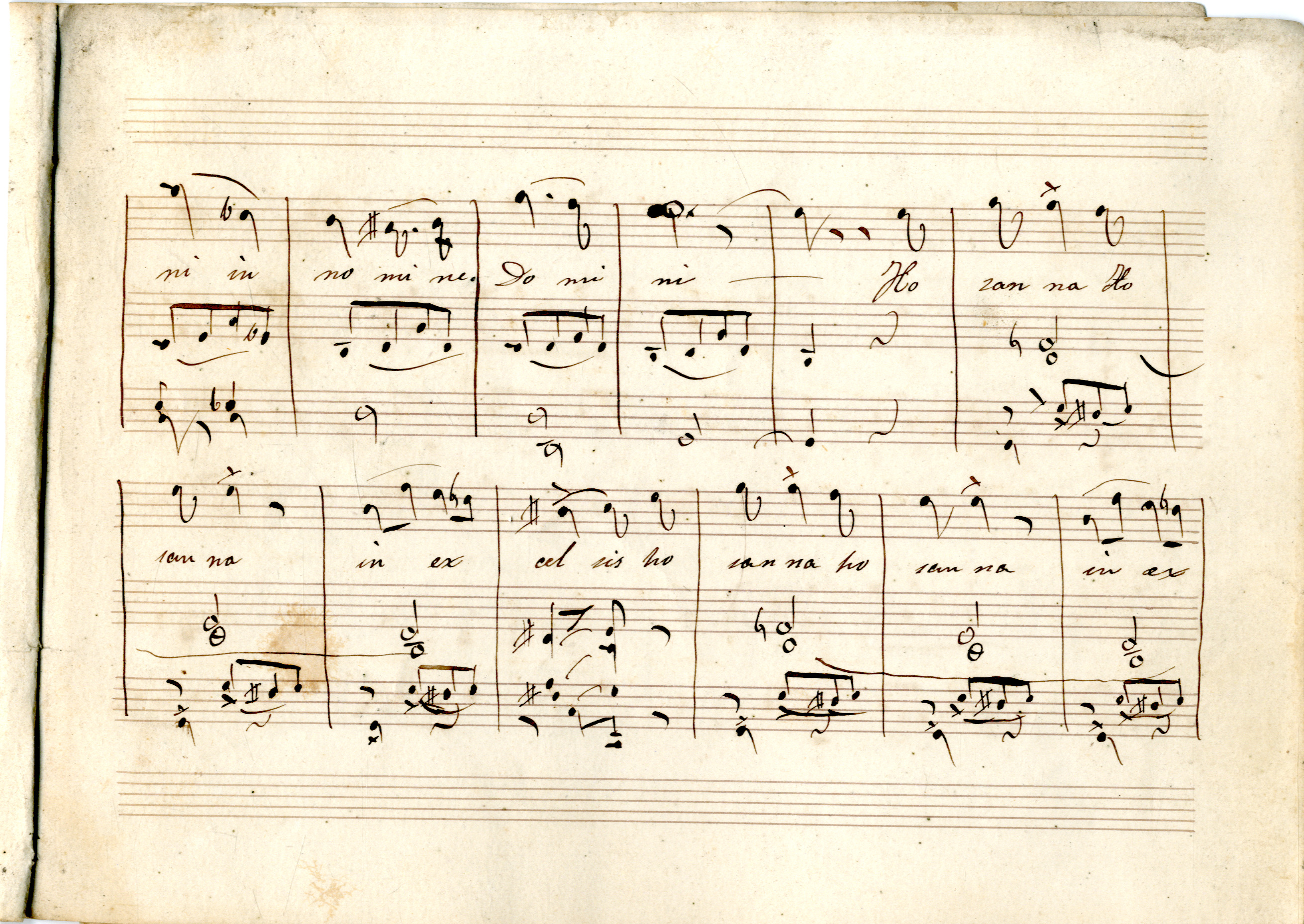}
    
    \caption{Examples of the diverse typologies of pages present in the dataset. There are orchestral works, particellas and pianoform scores, among others. Some examples are typeset and others handwritten, with diverse degrees of paper conservation quality.}
    \label{fig:dataset_examples}
\end{figure}

Optical Music Recognition (OMR) is the field devoted to the conversion of images of musical documents into computer-processable files \cite{calvo-zaragozaUnderstandingOpticalMusic2021}. As a field, it shares strong ties to other similar recognition tasks such as Optical Character Recognition (OCR, recognition of textual documents) or Graphics Recognition (e.g. floor-plans, engineering drawings). Nevertheless, OMR is considerably less developed as a technology than most of them, especially concerning historical and handwritten scores, despite the great strides that have been made in Artificial Intelligence as a whole in recent years.

There are many reasons for this situation; some of them are structural, such as the small community of researchers that are working full-time on OMR, the existence of limited economic incentives to develop it and the complexity of music notation itself. 
However, the biggest roadblocks that hinder progress in the discipline are technical \cite{torrasUnifiedRepresentationFramework2024}. In particular, there is a severe lack of transcribed scores readily available to train Deep Learning-based OMR systems. This limitation is even more acute in the context of handwritten or historical scores in Common Western Music Notation (CWMN), for which there are essentially no large-scale datasets at all, despite handwritten documents comprising a very large part of sheet music collections.\footnote{The central RISM catalogue of sheet music across libraries and archival institutions lists 1,348,727 manuscripts over just 261,537 prints (\url{https://rism.online/}, accessed 2026-03-10).} 
In practical terms this means that, as of today, there is no method that can reliably recognise these highly complex scores; the only attempts that exist still show very large error rates even for relatively simple scores \cite{baroHandwrittenHistoricalMusic2020}. 

Moreover, there is no established dataset for tracking progress toward practical applications, leaving a persistent gap between OMR experimental results and real‑world usability. Some works \cite{rios-vilaEndtoendOpticalMusic2023,mayerPracticalEndToEnd2023,rios-vilaEndtoEndFullPageOptical2026} show that current OMR methods have overcome model design difficulties inherent to the domain of music notation, and thus now have potential to achieve real-world impact. But unless the lack of realistic datasets is rectified, OMR cannot aspire to the proof-of-concept stage (Technology Readiness Level 4 \cite{lavinTechnologyReadinessLevelsML2022}). 

In order to address this gap, in this paper we introduce MusiCorpus, a dataset designed to advance Optical Music Recognition for historical and handwritten music scores. MusiCorpus is composed of 1,309 pages of hand-annotated historical and handwritten music, containing both MusicXML transcriptions at page and system level for end-to-end transcription, and harmonised primitive-level annotations for object detection-based methods. A sample of the pages included in the dataset can be seen in \autoref{fig:dataset_examples}. In this dataset, we also provide metadata for each of the musical documents, so that this collection can be used for other tasks such as writer identification and score dating in the future. Finally, we define structural constraints and evaluation metrics to encourage the OMR community to contribute data in a joint standardised schema --- in a similar fashion as Machine Translation's \textit{OpusCorpus} \cite{TIEDEMANN12.463}. This design is summarised in \autoref{fig:abstract}.

\subsection{Datasets on Music using the Common Western Music Notation}

\begin{figure}[ht]
    \centering
    \includegraphics[width=1.0\linewidth]{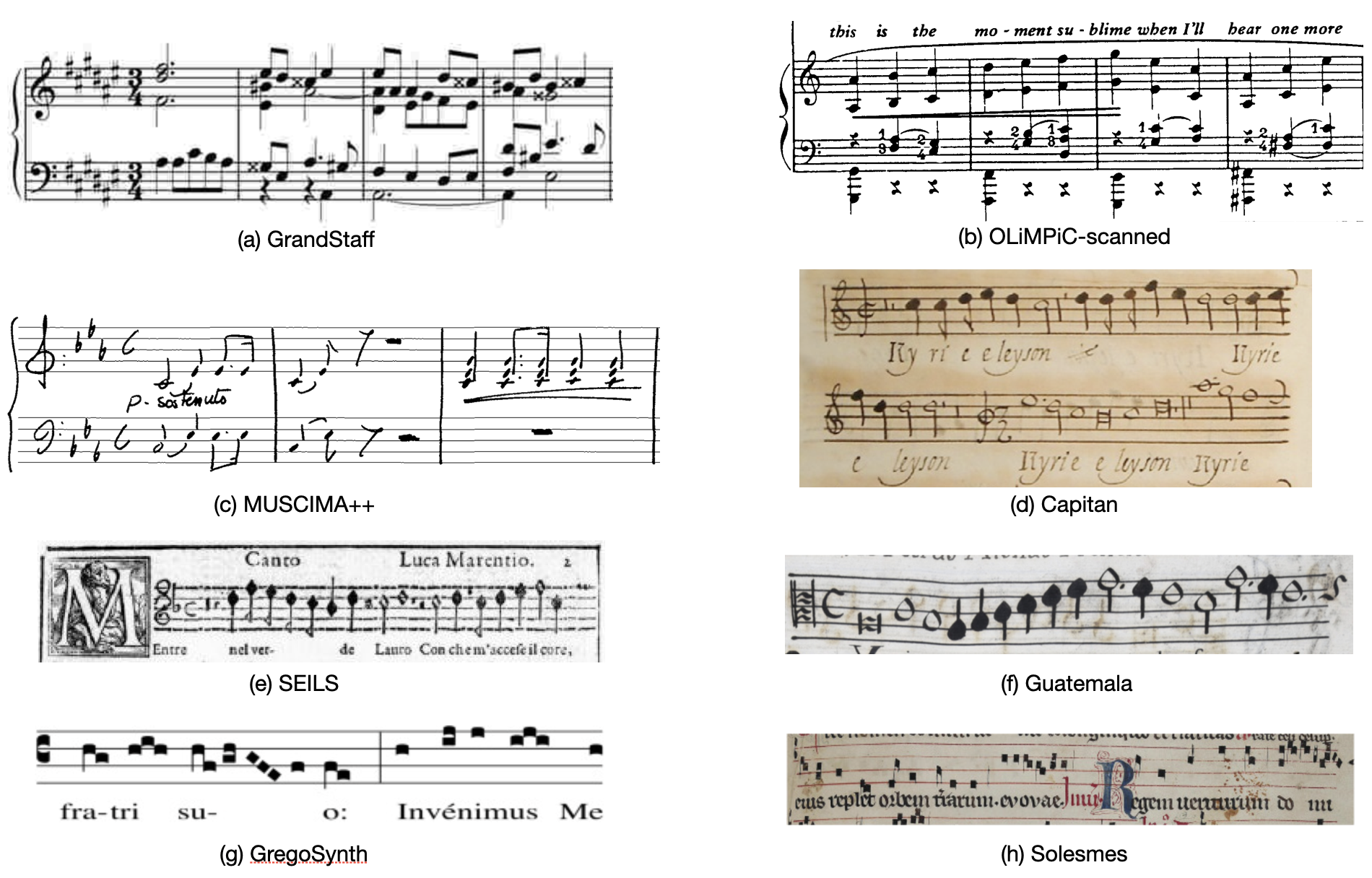}
    \caption{Illustrations of the types of music notation images contained in major OMR datasets. (a) Born-digital (synthetic) CWMN, (b) scanned image of CWMN, (c) handwritten and binarised CWMN collected for OMR experiments rather than taken from real collections, (d) and (f) mensural handwritten notation, (e) printed mensural notation, (g) born-digital (synthetic) chant notation, (h) real medieval manuscript of chant notation.}
    \label{fig:datasets-examples}
\end{figure}

As shown in \autoref{fig:datasets-examples}, there are multiple music notation systems, increasing the challenges in this field. In this section we overview the main music score datasets used for OMR using the Common Western Music Notation (CWMN).

\subsubsection{Born-digital images of printed notation}
Born-digital datasets comprise the largest fraction of datasets devoted to CWMN. These datasets have been created from digital renderings of music produced with engraving software such as MuseScore, Dorico, Verovio or Lilypond. This is the most straight-forward approach to OMR dataset creation, but the scores are considerably cleaner than real ones, which poses a significant limitation: models trained on these born-digital images often exhibit limited generalization when applied to practical real settings. Examples of born-digital datasets include:
The PrIMUS dataset~\cite{calvo-zaragozaEndtoEndNeuralOptical2018}, designed for end-to-end music recognition, extracted from the RISM database, and the Camera-PrIMUS dataset~\cite{calvo-zaragozaCameraPrIMuSNeuralEndtoEnd2018}, a variant that applies distortions to mimic scanned scores (e.g. blurring, rotation, etc.). The GrandStaff dataset~\cite{rios-vilaEndtoendOpticalMusic2023} consists of piano music extracted from the KernScores database~\footnote{\url{https://kern.ccarh.org/}} and rendered using Verovio \cite{puginVerovio}. Transcriptions use the \KERN notation format and the dataset is designed for end-to-end transcription. The GrandStaff dataset also has a Camera variant with similar distortions as the PrIMUS dataset.
The MSMD dataset~\cite{Dorfer2018MSMD} consists of piano music rendered directly from the Mutopia project\footnote{\url{https://www.mutopiaproject.org/}} and rendered using LilyPond \footnote{\url{https://lilypond.org/}}. It can be used for end-to-end recognition and also for music object detection thanks to the annotation of positions of noteheads in the images. Similarly, the DeepScores dataset~\cite{tuggenerDeepScoresDatasetSegmentation2018} and its V2 successor~\cite{tuggenerDeepScoresV2DatasetBenchmark2020} are also rendered from LilyPond. They are designed for music object detection, but do not contain ground truth for end-to-end recognition pipelines.
The OLiMPiC dataset~\cite{mayerPracticalEndToEnd2023} is derived from OpenScore Lieder~\cite{gothamScoresOfScore2018,gothamOpenScoreLieder2022}, a corpus of crowdsourced music transcriptions with permissive licenses. The dataset contains piano accompaniments of 19th-century songs, engraved using MuseScore. 
DoReMi~\cite{shatriDoReMiFirstGlance2021} is a dataset of 6,432 pages rendered with the Dorico notation software. At the time of its release it was one of the very few datasets to include annotations accommodating multiple OMR approaches. It provided ground truth for object detection, end-to-end recognition --- including MusicXML and MEI transcriptions --- and a graph-based annotation system in the spirit of MuNG, a format originally developed for the MUSCIMA++ dataset~\cite{hajicMUSCIMADatasetHandwritten2017}.

Concerning handwritten scores, it must be mentioned that Mashcima and Smashcima~\cite{mayerSmashcimaFullPageHandwritten2025} are both tools to synthesise handwritten-looking images from digital sheet music. They both leverage the symbol-level annotations found in datasets such as MUSCIMA++~\cite{hajicMUSCIMADatasetHandwritten2017} and compose them on a paper-like background to generate scores similar to those found in historical music collections. Smashcima in particular can render a full score from a MusicXML file and also provide bounding box annotations, which allows it to produce datasets that support both end-to-end and object detection pipelines.

\subsubsection{Scans of printed Scores} 
The next rung in the difficulty ladder are printed scores where the digital artifact is not rendered directly from an encoding such as MusicXML, but produced with imaging technology --- scanners or cameras. This means the resulting documents contain alterations caused by the imaging process such as scanning post-processing, lighting condition changes, misalignment and imperfect flattening of bent pages, as well as the damage present on the physical support of the documents. 
The OLiMPiC-Scanned dataset~\cite{mayerPracticalEndToEnd2023} is a subset of the OLiMPiC dataset for which images of individual piano staves from scans hosted in the IMSLP online sheet music library are provided. It contains a total of 2,931 annotated samples. The ground truth, as with the original OLiMPiC dataset, is annotated in MusicXML, suitable for end-to-end recognition.
The CollabScore corpus~\cite{rigaux:hal-05515751}
contains 26 scores by Camille Saint-Saëns, totaling 199 notated pages or 649 systems. The dataset has annotations in MEI and MusicXML formats created from manual transcriptions using notation software and is tailored to end-to-end recognition. Within the constraints of late 19th and early 20th c. music printing, the pages are visually varied.

\subsubsection{Handwritten scores} 
Several datasets with handwritten music notation exist, but they are either small in number of samples or limited in what ground truth they provide. Creating ground truth for handwritten scores, especially for object detection, is very time-consuming and expensive, since the annotation process is usually fully manual, the documents tend to be very dense with symbols and the people who perform this process need to be highly specialised.

The CVC-MUSCIMA dataset~\cite{fornesCVCMUSCIMAGroundTruth2012} contains 1,000 binarised images of 20 pages of music copied by hand by 50 different writers. These images are then altered using 11 different distortion algorithms, producing a total of 12,000 images when combined with the unmodified ones. This dataset, however, was not designed for transcription tasks, instead being focused on staff removal~\cite{gallegoStaffLineRemoval2017,dvorakStaffLayoutAnalysis2024} and also writer identification. As a matter of fact, it does not contain any transcription information at all.

The MUSCIMA++ dataset~\cite{hajicMUSCIMADatasetHandwritten2017} addresses part of this problem by providing object detection ground truth for 140 of the 1000 pages of CVC-MUSCIMA. It also pioneered the Music Notation Graph (MuNG) format, which allows to go beyond object detection to extract musical semantics from the page by combining these detections using a graph structure. However, the dataset does not provide complete transcriptions in MusicXML or some other high-level encoding, so it is useful only for object detection-based pipelines. It also inherits the binarised nature from the original CVC-MUSCIMA images, so despite being handwritten, it still generalises poorly to real-world scores.

The Pau Llinàs dataset~\cite{baroOpticalMusicRecognition2019} is a very small (246 samples) collection of measure-level images transcribed in an agnostic notation format extracted from an 18th century Motet by the homonymous Catalan composer. It is one of the few available datasets for end-to-end recognition of handwritten historical scores. Similarly, there is the FMT database~\cite{martinez-sevillaTowardsUniversalOptical2024}, which contains real scores --- including manuscripts --- collected from 20th-century ethnomusicological records of the Fondo de Música Tradicional IMF-CSIC~\footnote{https://musicatradicional.imf.csic.es/}. However, this dataset is limited to monophonic music, and to the best of our knowledge, it is not publicly available.

The HOMUS collection~\cite{calvo-zaragozaRecognitionPenBasedMusic2014}  is a dataset of music symbols captured using a tablet, which makes it one of the few datasets with online handwriting information on the list. It contains 15,200 instances of 32 unique musical symbols written by 100 different musicians. It can be used for offline classification of music symbols by rasterising the stroke information.

\subsection{Music in Mensural and other Notation Systems}

While not directly related to this work, it should be noted that there are important datasets built around other musical traditions and notation systems than CWMN.
A big portion of OMR research has historically been focused on mensural notation works, which is reflected in the considerable number of datasets available for this specific notation system. Most of the datasets have originated from scans of specific volumes, such as SEILS~\cite{parada-cabaleiroSEILSDatasetSymbolically2017}, Mottecta~\cite{martinez-sevillaPerformanceOpticalMusic2023}, Capitan~\cite{calvo-zaragozaTwoNoteHeads2016} or Guatemala~\cite{10.1145/3543882.3543885} collections. Many of these volumes are actually handwritten manuscripts, contrary to the general trend of CWMN recognition. There have also been attempts at synthesising documents in Mensural notation, as seen in PrIMens~\cite{martinez-sevillaPerformanceOpticalMusic2023}.

Similarly, significant attention has been devoted to Gregorian chant notations. Gregorian chant was the overwhelmingly most important musical tradition in medieval Europe with over 30,000 extant manuscripts; for this era of European music it is essential. Recently, four datasets have finally been released: two datasets from individual manuscripts, Salzinnes and Einsiedeln, a mixed dataset, Solesmes, and a synthetic dataset, GregoSynth \cite{fuentes-martinezAlignedMusicNotation2026}. However, both mensural and Gregorian notation only allow for monophonic (single-voice) music, and the notation styles --- while re-using some essential elements such as staffs, noteheads, and stems --- are significantly different.

Outside of European notations that use the staff, some OMR datasets have been created for Jazz lead sheets \cite{martinez-sevillaOpticalMusicRecognition2025}, the KuiSCIMA dataset of traditional Chinese Suzipu notation \cite{repoluskKuiSCIMADatasetOptical2024}, and a dataset is available also for Korean Jeongganbo notation \cite{kimAutomaticRecognitionOfJeongganbo2025}.


\section{Methods}

The dataset is designed with supervised machine learning for Optical Music Recognition as its primary goal. At its core, it contains pairs of music document scans and their transcriptions. However, music notation is not as straightforward to transcribe as similar domains such as text, as there are many valid approaches --- each with their tradeoffs --- both to the process of encoding and recognition of musical sources. Moreover, we anticipate that a collection of musical documents such as this one can potentially be used for more tasks than initially intended if some minimal archival information for each of the images is also provided as metadata.

The dataset has been created across two data collection sites: Dolores\footnote{https://pages.cvc.uab.es/musicscores/} (\textbf{Do}cument Understanding in \textbf{Lo}w-\textbf{Res}ource Scenarios), and OmniOMR\footnote{https://ufal.mff.cuni.cz/grants/omniomr}. Each site collaborated with a local memory institution to obtain representative samples of sheet music in their collections. The OmniOMR site worked with the Moravian Library\footnote{\url{https://www.mzk.cz/en}}, while the Dolores site collaborated with the Xarxa d'Arxius Comarcals de Catalunya (XAC), the Societat del Gran Teatre del Liceu, the Institució Milà i Fontanals of the Spanish National Research Council (IMF-CSIC), and the Documentation Centre of the Palau de la Música Catalana (CEDOC). These institutions have generously allowed publishing the original material under a Creative Commons Non-Commercial Share-Alike license.
Experts from these institutions also provided metadata for the selected page images.

Thus, in \autoref{subsec:transcription} we first describe the choices we made for transcription formats in this dataset, contextualising them with the recognition approaches they try to enable, whereas in \autoref{subsec:dolores} and \autoref{subsec:omniomr} the annotation methods for each of the dataset parts corresponding to the individual data collection sites are presented.

\subsection{Ground truth formats} 
\label{subsec:transcription}

There are two main groups of OMR methods: end-to-end recognition~\cite{calvo-zaragozaEndtoEndNeuralOptical2018,baroOpticalMusicRecognition2019,baroHandwrittenHistoricalMusic2020,torrasIntegrationLanguageModels2021a,rios-vilaEndtoendOpticalMusic2023,fuentes-martinezAlignedMusicNotation2024,garrido-munozHolisticApproachImagetograph2022,rios-vilaEndToEndFullPageOptical2022,rios-vilaEndtoEndFullPageOptical2026}, and object detection-based pipelines \cite{lemaitre_collabscore_2026,hajicTowardsFullPipeline2018,pachaLearningNotationGraph2019}.
The former has recently been much more prevalent for Western notations (CWMN, mensural, chant), while the latter has been used for non-staff notations (Jeongganbo, Suzipu) but has been competitive for western notations previously, and has interpretability advantages that translate to easier maintenance and fine-tuning of deployed systems. 
No head-to-head comparison of these approaches has been performed, in part, because each approach requires a different type of ground truth.
We provide ground truth for both approaches, also to enable this head-to-head comparison.

The end-to-end approach requires only the target encoding of music notation. The detection-based pipeline requires annotating the individual notation symbols in the image. Then, in order to construct the target encoding from individual symbols, one must relate the individual notation symbols to each other \cite{rebeloOMRStateOfTheArt2012,hajicMUSCIMADatasetHandwritten2017,calvo-zaragozaUnderstandingOpticalMusic2021}, for which a graph or a graph-like structure (such as a parse tree) is needed \cite{hajicMUSCIMADatasetHandwritten2017,couasnonDMOSGenericDocument2006,torrasUnifiedRepresentationFramework2024}. However, the main machine learning challenge, and the main requirement for training data, still lies in the object detection phase.

\subsubsection{End-to-end ground truth.} 

Multiple structured representations of music notation exist with sufficient tooling available: MusicXML, \KERN, MEI, LilyPond and ABC, among others, but none of them stands out as a standard for the OMR community. Outside of OMR, MEI and MusicXML seem to be the most widely used notation systems and are routinely employed by musicians and musicologists. MusicXML is likely the best-supported music format: it is the format that has the most tools, converters and scores readily available, and has the most support among widely used music notation editors (MuseScore, Sibelius, Dorico). It is the only format, in fact, that has conversion available to and from (most of) the others somewhat reliably. Therefore, the choice of MusicXML imposes the fewest limitations on OMR system development: a system can convert MusicXML encodings to the format that works best for the method in question, convert results to MusicXML, and then evaluate the results using a normalised MusicXML metric~\cite{knopkeMusicdiffFoundationImproved2007,torrasUnifiedRepresentationFramework2024}. Other attempts at building benchmarks have taken a similar route \cite{byrdStandardTestbedOptical2015,hajicFurtherSteps2016,torrasUnifiedRepresentationFramework2024,martinez-sevillaSheetMusicBenchmark2025b}.

\subsubsection{Detection-based ground truth: Two strategies.} 

Ground truth for object detection is provided in the standard MSCOCO~\cite{linMicrosoftCOCOCommon2014} format.
For object class names we adhere to the SMuFL standard as much as possible, to ensure maximal compatibility with the rest of the music notation ecosystem. In those cases where SMuFL does not define a specific symbol (e.g. beams), name definitions are taken from the MuNG format instead \cite{hajicMUSCIMADatasetHandwritten2017}.

At this junction, an important decision should be made. Given the fact that the annotation process for object detection ground truth is manual, there is a tradeoff between accuracy and scale given a fixed amount of available work hours for annotation. One can either annotate with very high accuracy and accept slow progress, or prefer faster progress at the cost of providing somewhat noisy ground truth. Each annotation strategy also comes with different needs in tooling and process control. In this work, we decided to explore opposing ends of this decision space while producing a single, larger dataset:

\begin{itemize}[noitemsep]
    \item On the one hand, a pixel-perfect symbol mask annotation and full notation graph (MuNG) strategy was implemented at the \textbf{OmniOMR} site, yielding 100 pages in total.
    \item On the other hand, a fast and coarse stylus-based strategy was implemented at the \textbf{Dolores} site, producing 1,209 pages instead.
\end{itemize}

Both resulting collections are compatible and standardised, powering different use cases when used separately but enabling the standard OMR pipelines when used jointly. Moreover, we hypothesise that the existence of the highly-precise OmniOMR dataset could enable refinement of the Dolores annotations in the future.

\subsection{Data selection and collection procedure: Dolores site} 
\label{subsec:dolores}
\begin{figure}[ht]
    \centering
    \includegraphics[width=\linewidth]{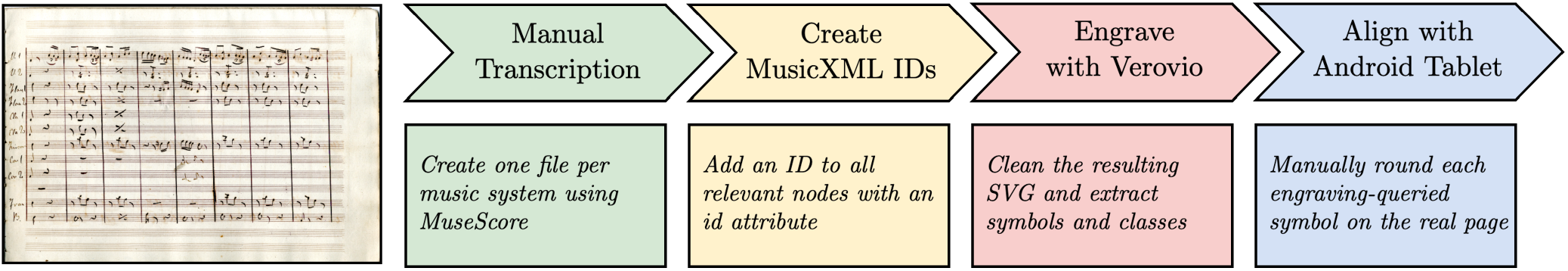}
    \caption{Overview of the Dolores annotation pipeline.}
    \label{fig:dolores-pipeline}
\end{figure}
The decisions that were taken in developing this fraction of the dataset try to balance the desire of annotating the maximum number of scores possible while still having primitive-level annotations and transcriptions, trading off primitive-level annotation accuracy for scale. A summary of the overall process can be found in \autoref{fig:dolores-pipeline}.

\subsubsection{Data Selection}
The subset of Catalan musical works was selected according to various criteria, both musicological and technical. From a musicological perspective, and given that the methodology proposed in the project required a substantial amount of material, the intention from the outset was to construct a corpus that would be both diverse and representative of CWMN.

With this aim in mind, the selection process was carried out jointly by technical experts, musicologists, and the institutions that safeguard the original documents and that made their digitised collections available for the project. This active and deliberate involvement of the main institutions dedicated to the preservation and management of Catalan musical heritage made it possible to propose a corpus consisting predominantly of works by Catalan composers. The overall chronological framework, covering compositions from 1700 to 1950, was selected to correspond to the central period in the development of Common Western Music Notation (CWMN), during which only minor variations in notational symbols were used.
%
\subsubsection{Annotation}
To get the transcription for each page, we could not find any faster alternative than manually copying the source material using MuseScore, a tool most musicologists were fairly proficient with before joining the project. The pages are transcribed as literally as possible, restricting the editorial modifications to those contexts where engraving would have been impossible otherwise. To speed up the process, we chose not to annotate directives, lyrics, expression slurs and other expression indications. 

We made the choice of annotating each System of music separately to facilitate posterior processing, anticipating that merging MusicXML files would be easier than splitting them. Our definition of a music System corresponds to all staves that contain parts to be played in unison, which means that each of these transcription files can contain multiple parts for multi-instrument pieces and there can be more than one of these files per page. This is exemplified in \autoref{fig:layout-definition}. All resulting MuseScore files are batch-converted to MusicXML, modified to incorporate useful symbol-level identifiers and merged together to obtain a singular page-level file that is compliant with OpenScore directives. This page-level file does not retain the symbol-level identifiers due to technical limitations ---~we merge via Music21\footnote{\url{https://music21.org/music21docs/index.html}}, which performs an intermediate conversion to their own internal format~--- so for symbol-level processing we restrict ourselves to using the system-level files.

The idea behind the aforementioned primitive identifiers provided on the System-level MusicXML files is that the visual elements of each score can be tied to their semantics. Unfortunately, MusicXML does not provide a way of giving identifiers to \textit{all} elements in the XML tree, nor does it represent every visible symbol as a separate XML element --- \eg \texttt{notehead}s are optional and do not have an \texttt{id} attribute, or key accidentals can be represented by the number of fifths of the key and not as a list of accidentals, each in its corresponding position. We solve this issue by constructing new identifiers derived from their parent elements. For the example of noteheads, the id \texttt{note<N>.notehead} is the identifier of the notehead associated to the note node with id \texttt{note<N>}. Many times only one instance of a child element is allowed, in which case we do not need to add any further disambiguation information to the identifier. In those cases where more than one object of the same class is allowed to be a child of an existing element, we number them sequentially in the order of appearance on the engraving. Most of these child elements tend not to have semantics tied to their specific order of appearance; for those where order matters (most prominently, barlines), we rely on the fact that our engraver is consistent in the order it generates elements in the graphical representation of the score.

\begin{figure}
    \centering
    \includegraphics[width=0.46\linewidth]{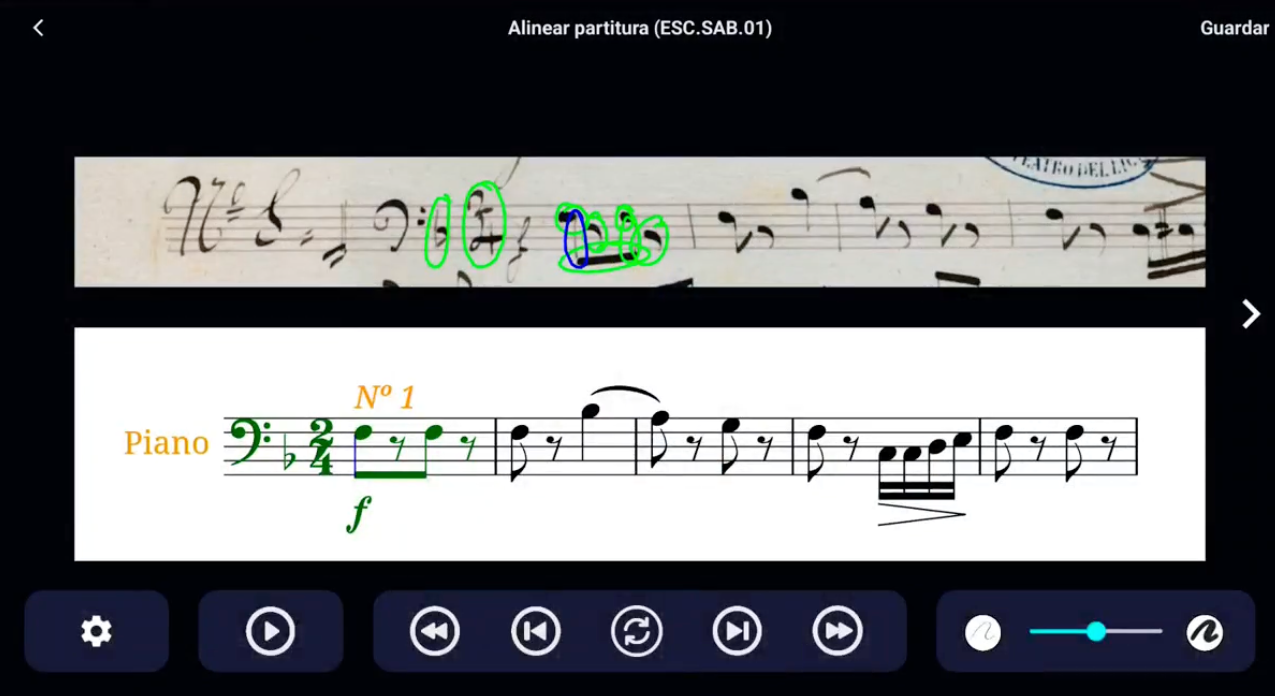}
    \includegraphics[width=0.5\linewidth]{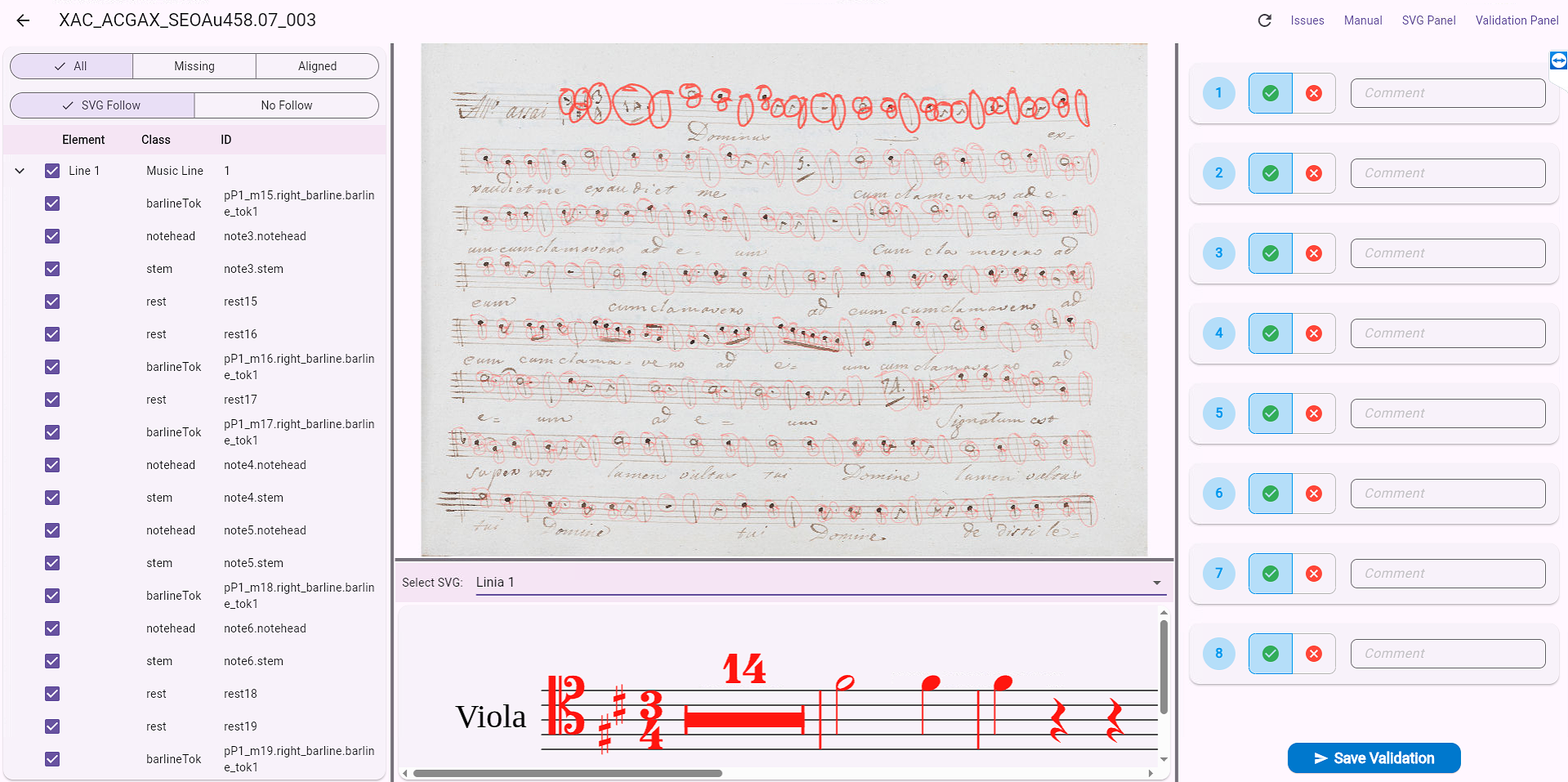}
    \caption{Showcase of the various tools used by the Dolores site. On the left, a screenshot of the Android app used by the musicologists to annotate the symbols on the transcribed score over the original page. On the right, a screenshot of the validation tool used to quickly verify that all annotations are present and follow the desired format, as well as an example of a fully annotated and validated page using the Dolores spec. These are both connected and supported by the \texttt{lola} backend application that handles storing the annotations and processing all of the involved files.}
    \label{fig:dolores-tools}
\end{figure}

Once the manual MusicXML transcriptions are obtained, a tool-assisted pipeline is deployed in order for musicologists to manually annotate each primitive on the page and tie it to an object in the MusicXML file, using the identifiers that were generated during the previous step. We have developed an Android application, shown in \autoref{fig:dolores-tools}, that allows musicologists to identify objects on the original music score page by coarsely drawing circles around them using a stylus pen. The application prompts them to identify primitives by changing the colour of the symbol in question in an engraving of the transcript of the score. After a symbol is identified, the application automatically saves the drawn polygon, the identifier of the object and its class and proceeds to the next symbol without any further input from the musicologist. For this process, we leverage implementation details from the Verovio engraver that allow us to process individual primitives on the score and get their MusicXML identifiers and their SMuFL class \footnote{Incidentally, we have also found it possible to automatically generate datasets that are MusiCorpus-compliant from digital-born scores using this pipeline.}.

\subsection{Data selection and collection procedure: the OmniOMR site}
\label{subsec:omniomr}

The OmniOMR portion of the data attempts to produce a collection of samples that is as accurate as possible in order to enable a broader set of downstream tasks. It provides pixel-accurate symbol masks at the cost of having steeply increased annotation requirements, thus resulting in a much smaller dataset. However, it is more diverse in terms of the musical and visual styles therein. 

\subsubsection{Data Selection}

The musical documents for the OmniOMR site have been selected by curators of the musical collections of the Moravian Library.
The high level of detail of the strategy implemented in this part of the dataset means that much fewer pages could be annotated, necessitating in turn a different data selection policy to maximise visual and musical diversity. Hence, the OmniOMR data does not contain entire works but rather individual pages. At the same time, however, to enable some experiments with adaptation to handwriting styles, we choose on average 2 pages from each source.

\subsubsection{Annotation} 

\begin{figure}
    \centering
    \includegraphics[width=0.7\linewidth]{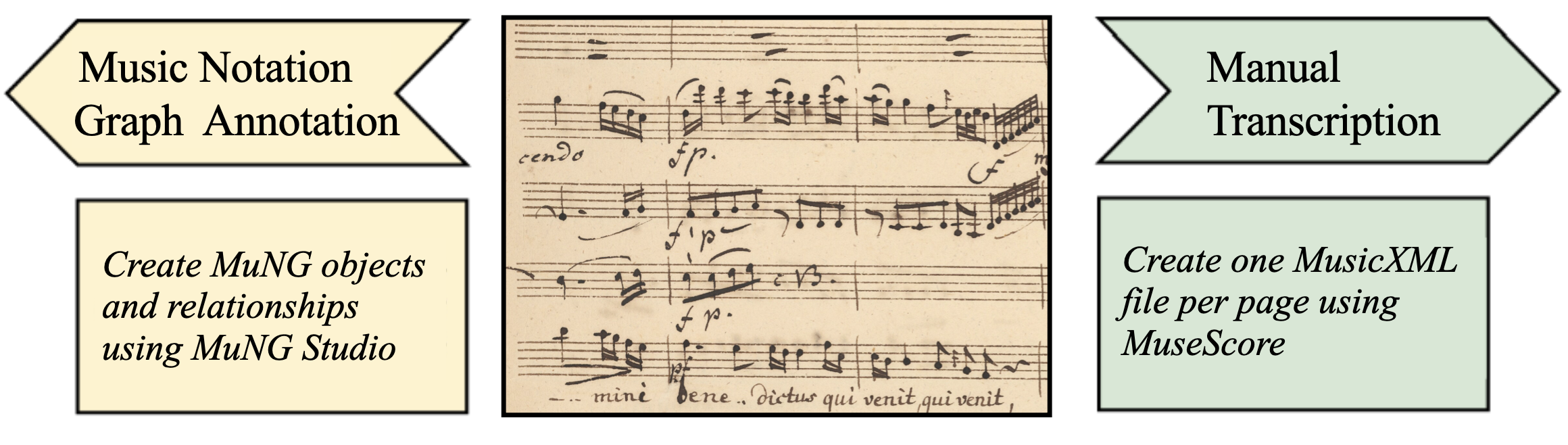}
    \caption{OmniOMR site annotation schema: separate tooling for MusicXML transcription and highly accurate symbol annotation, including the full MuNG standard, using MuNG Studio. Revisions (not shown) were done for both outputs, as well as cross-checks against each other.}
    \label{fig:omniomr-annotation-process}
\end{figure}

A summary of the OmniOMR annotation process is shown in \autoref{fig:omniomr-annotation-process}. The MusicXML transcriptions were performed using the \mbox{MuseScore} notation software following the methodology of the OpenScore corpus \cite{gothamScoresOfScore2018,gothamOpenScoreLieder2022}. All annotators had musical education and prior experience with using MuseScore. The role of editors responsible for checking correctness was fulfilled by members of the core team from the Moravian Library (who also have extensive experience with the notation software). Quality control included both visual inspection and replaying the encoded music from the editor, mainly to find issues with accidentals and key signatures that are difficult to spot by eye.

Data quality was further checked with automated tests against issues with assigning individual staffs to systems and grand staffs, which are invisible both to the editor's eye and to the replay functionality of MuseScore. In 16 pages the MusicXML format did not match the verified MuNG annotations, so they had to be manually fixed.


The Music Notation Graph (MuNG) standard \cite{hajicMUSCIMADatasetHandwritten2017} consists of two types of elements that must be manually annotated: symbols (nodes of the notation graph), and their relationships (edges of the notation graph).
The format was created with machine learning in mind: even though musical semantics are non-local (\eg the pitch of a note depends on a combination of symbols in a different part of the image), detecting objects and classifying edges can be done within the field of view of a reasonable stack of convolutional filters (\eg 200 pixels of the input image).

Symbols in the OmniOMR data are annotated with a pixel-accurate mask.
Object type names are taken from the SMuFL specification for music notation glyphs when possible. Otherwise, they are taken from the MUSCIMA++ dataset, which is the previous dataset that uses MuNG (\eg \texttt{beam}).

MuNG edges have two types: \textit{syntax}, which links together the visual symbols that must be interpreted as a group, and \textit{precedence}, which define the order in which the music should be read. They are added by annotators as annotation progresses.

During the first iteration of the annotation process, multiple deficiencies of the original MuNG format were discovered and the MuNG 2.0 standard was created.\footnote{The definition of the MuNG 2.0 format is given in \url{https://github.com/OmniOMR/mung/blob/main/docs/annotation-instructions/annotation-instructions.md}.}

\begin{figure}
    \centering
    \includegraphics[width=1.0\linewidth]{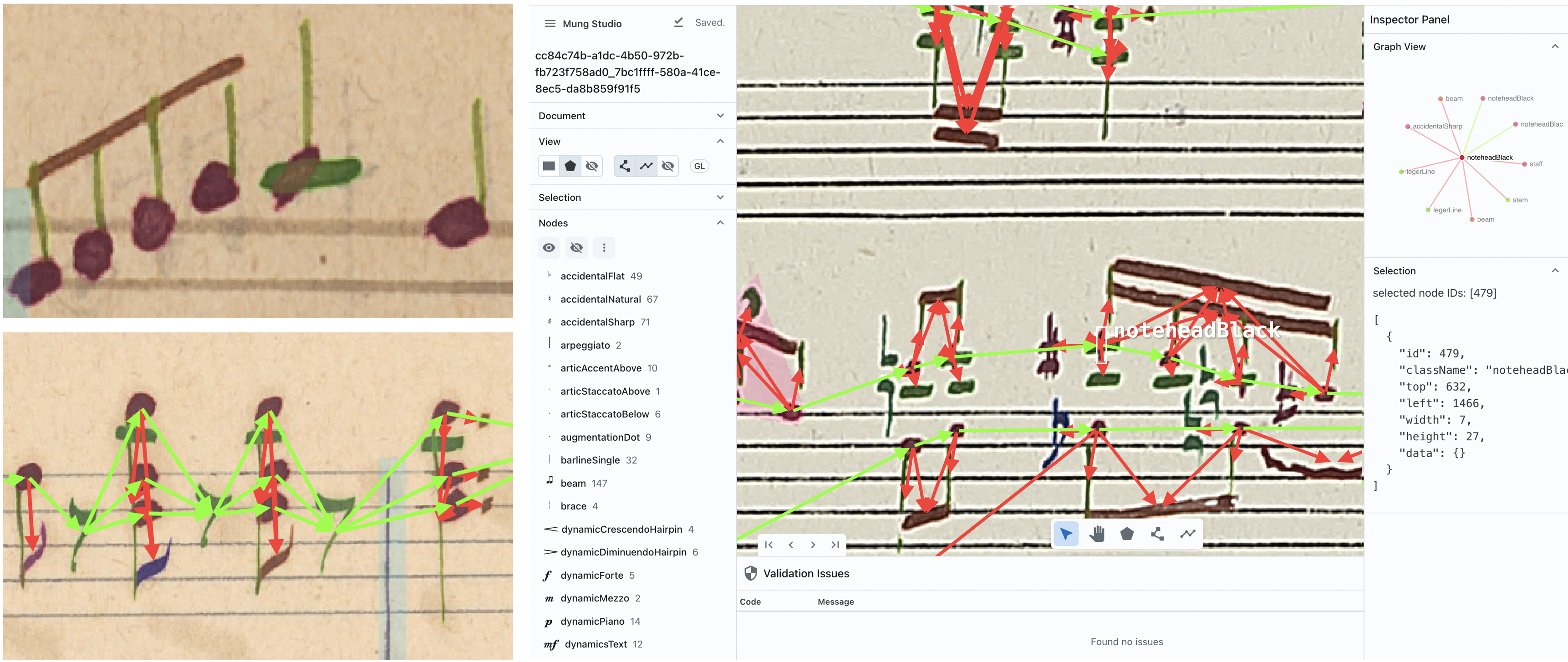}
    \caption{Showcase of the annotations by the OmniOMR site. \textbf{Top left:} high-accuracy symbol annotations; a semi-transparent overlay indicates a each symbol class with color: noteheads purple, stems yellow, beam orange, etc. \textbf{Bottom left:} syntax and precedence edges of the MuNG format. \textbf{Right:} the MuNG Studio annotation web application, with inspection and validation tools.}
    \label{fig:omniomr-tooling}
\end{figure}

The MuNG annotation was performed using the MuNG studio tool \cite{mayerMuNGStudioAnnotation2025}, which can be seen in \autoref{fig:omniomr-tooling}. Annotators were trained over a 6-month period, and then cross-trained also to revise each other's work, so that each MuNG file went through a revision process. Automatic validation was implemented in MuNG Studio to catch frequent errors (\eg a syntactic edge in the opposite direction) and to pass the output of revisions.

\subsubsection{Train-Dev-Test Splits}

Given the diversity of the OmniOMR data while only containing 100 pages, multi-criterial stratification across notation metadata fields with controlled vocabularies (notation type, notation complexity, production, clarity, and distribution into systems) was performed. To evaluate a proposed split, for each metadata field, the categorical distribution of its possible values was computed across the train, test, and dev portion, and Euclidean distances between each pair of portions were summed. These values were then summed over all the controlled-vocabulary metadata fields. Because of the low number of images, to find an optimally stratified split, it was possible to try $10^6$ splits (uniformly over page permutations) and choose the best-performing one. After a manual inspection of the resulting optimal stratification (aiming at a difference of at most 1 between the smaller validation and training splits as the manually verifiable criterion for stratification quality), this split is then fixed.

The same stratification strategy was run once without accounting for multiple pages being from the same document, and once in a regime where all pages from the same document were only allowed to reside in one of the train/dev/test portions of the split. This policy ensures the specific scribal hands from the training data are never seen in dev and test, thus ensuring a higher standard for generalisation; however, we also retain the less strict split to simulate a situation where an experimenter provides a small amount of in-domain annotations for recognising a specific manuscript (which is one of the main workflows in OMR \cite{fujinagaArtOfTeachingComputersSIMSSA2019,harteltOpticalMedievalMusicRecognition2024}).

\section{Data Records}

\begin{figure}
    \centering
    \includegraphics[width=1.0\linewidth]{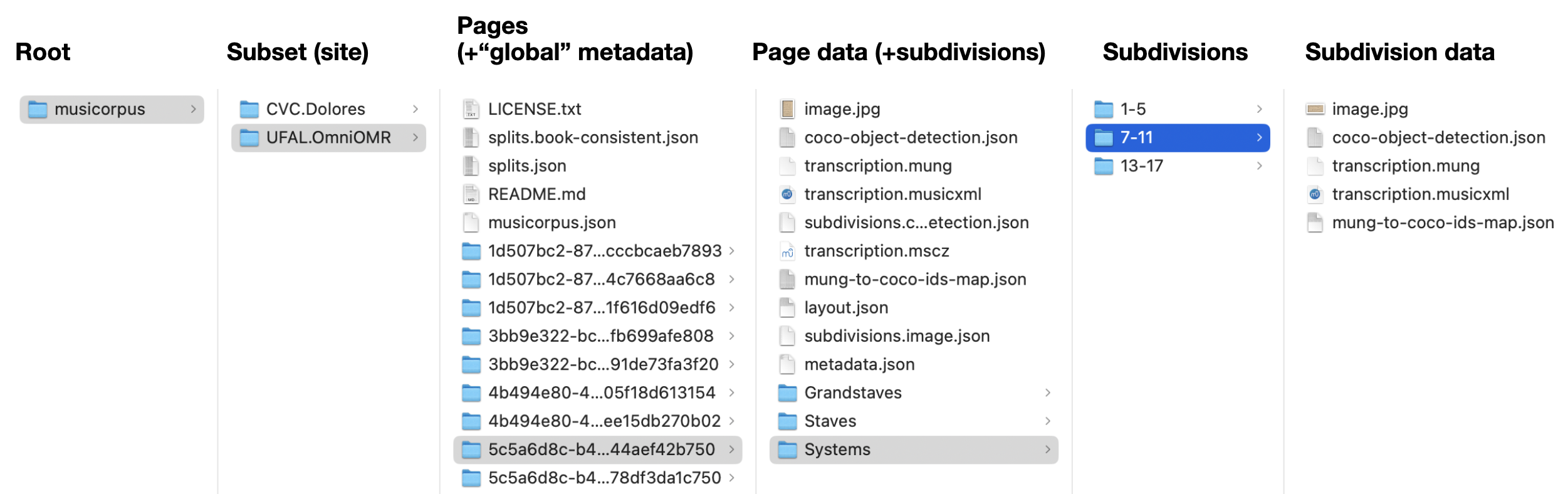}
    \caption{Directory structure of Musicorpus: site- (subset-)level directories, page-level directories with primary data, and subdivision-level directories with corresponding subsets of data for each page.}
    \label{fig:data-record-directory-structure}
\end{figure}

The data records for MusiCorpus are complex entities, reflecting the hierarchical decomposition of sheet music pages into systems and staffs, and the multiplicity of ground truth formats provided. The structure of the dataset from the root down to the subdivision level is shown in \autoref{fig:data-record-directory-structure}.

The full detailed documentation for the data records is available as Appendix A; in this section, we describe the fundamentals.

\subsection{Directory structure} 
The top-level items in each site's subset are \textbf{page} directories. The page ID is the directory name. The combination of subset name (\texttt{CVC.Dolores} and \texttt{UFAL.OmniOMR}) and page ID is guaranteed to be unique.
Each page then contains subdivisions: at the level of \textbf{Systems}, \textbf{Staves} and \textbf{Grandstaves}. Subdivisions are provided because many end-to-end OMR systems train at the system and/or staff level (full-page OMR is very recent \cite{rios-vilaEndtoEndFullPageOptical2026}), and splitting pages into subdivisions or combining subdivisions into pages is highly non-trivial; hence it is prudent to provide subdivisions as static entities. Page directories and individual subdivision directories then contain the primary data.
\subsection{Primary data}
For each page, the score image is available as \texttt{image.jpg}, alongside the two core ground truth files for the dataset: a MusicXML file named \texttt{transcription.musicxml} for page-level end-to-end recognition and an MSCOCO-format file named \texttt{coco-object-detection.json} containing the symbol-level annotations for the page. Datasets adhering to this standard can then optionally incorporate files in any other format their designers require to provide additional sources of supervision. In MusiCorpus, the files from the OmniOMR portion of the dataset also include a \texttt{transcription.mung} file containing notation assembly in MuNG format and a mapping file between MSCOCO and MuNG object identifiers named \texttt{mung-to-coco-ids-map.json}.

Optionally, the \texttt{layout.json} file specifies the bounding boxes of individual sheet music layout elements if they are provided in the dataset: systems, staves (in the sense of just the sets of parallel lines) and grand staffs (piano-like paired staffs), systems, and measures (at each subdivision level: staff measures, grand staff measures, and system measures).
\subsection{Subdivisions}
The structure of the page-level directory is repeated for each subdivision (System, Staff, and Grandstaff): each subdivision directory contains its own \texttt{image.jpg}, \texttt{transcription.musicxml}, \texttt{coco-object-detection.json}, etc. files. In code, the same loader can therefore be used to load data at any subdivision level.

\begin{figure}
    \centering
    \includegraphics[width=0.8\linewidth]{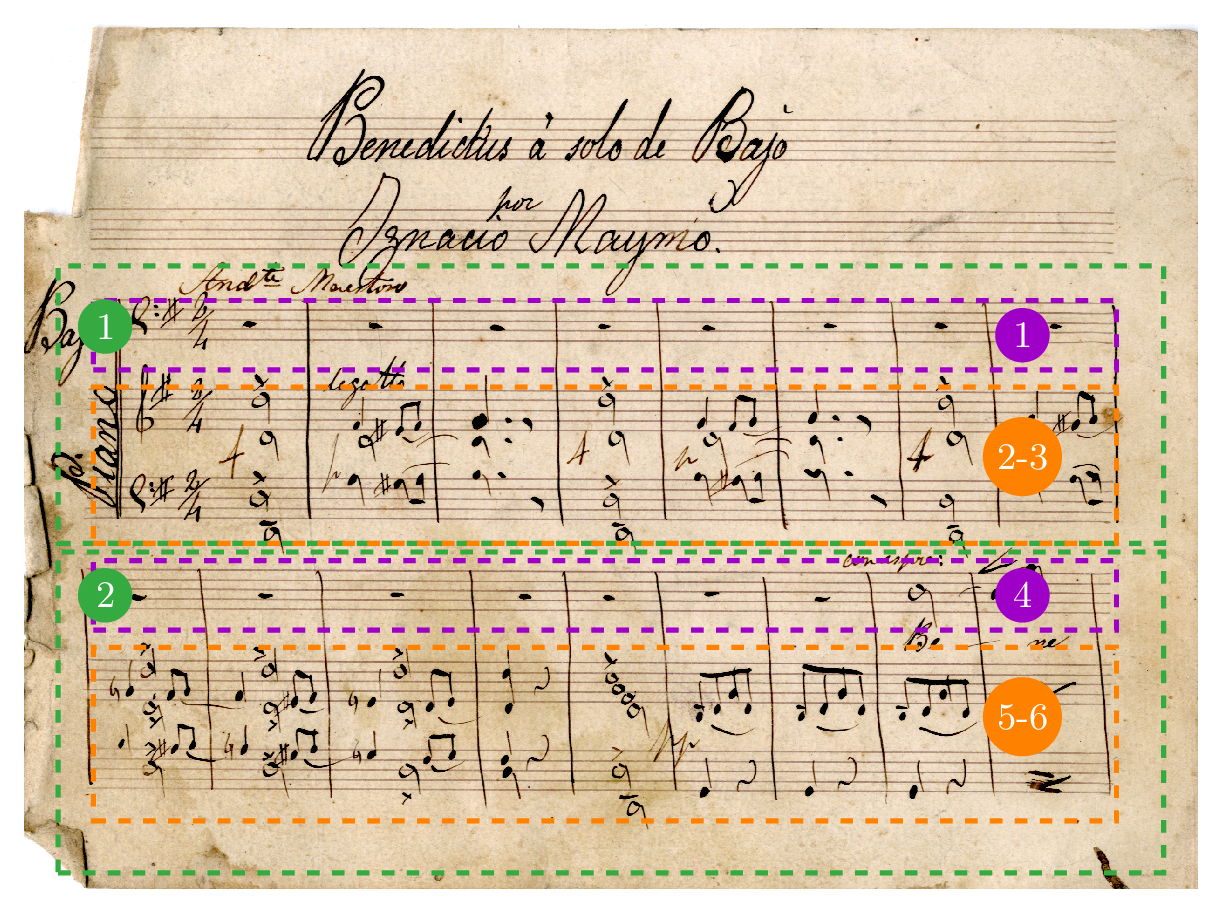}
    \caption{Definition of subdivisions within the dataset using a heterogeneous page as example. In green, the definition of Systems 1 and 2, which corresponds to sets of music that sound in unison. In purple, the definition of Staves 1 and 4, which correspond to lines of music written for instruments that require a single staff. In orange, the definition of Grandstaves 2-3 and 5-6, which are blocks of multi-staff music instruments.}
    \label{fig:layout-definition}
\end{figure}

Additionally, because the subdivisions relate to the full page, mappings between the page-level and subdivision-level information are provided in the page folder. To specify how the subdivision images relate to the full page image, the \texttt{subdivisions.image.json} file specifies the bounding box of the original page image that was used to crop out the subdivision's \texttt{image.jpg}. This differs from how layout elements are represented in the page-level \texttt{layout.json} file, where their ``tight'' bounding boxes are given, while here the ``loose'' bounding boxes that also contain all music notation related to that particular staff is included.
To specify how the object detection ground truth is related between the page level and subdivision level, a \texttt{subdivisions.coco-object-detection.json} file defines which page-level MSCOCO object ID maps to which subdivision MSCOCO object ID. The relationship between each of the subdivisions and the page is demonstrated in \autoref{fig:layout-definition}.

\subsection{Dataset Blocks}
The Dolores and OmniOMR portions of the MusiCorpus dataset each have their own directory, with the same page/system structure. This structure allows for clearly recording the origin of individual pages, which matters in the context of re-using archival materials, as they may each be licensed separately. Furthermore, this structure allows for other datasets to be integrated into the same compatible standard without sacrificing their identity, which is important to potential contributing institutions. The fact that a directory contains data in MusiCorpus-compatible format is signalled by the presence of a \texttt{musicorpus.json} file in its root directory, which specifies metadata such as the originating institution name, URL, timestamp, data versioning, and contact information for maintainers of the given dataset portion.

Within the root directory of each portion of the dataset are further descriptive files. A \texttt{splits.json} file specifies the canonical split into training, validation, and test pages for the dataset. For the OmniOMR portion, we also provide a split that assigns both pages from each source into the same train, test, or validation subset (file \texttt{splits.book-consistent.json}), a \texttt{README.md} that describes the specifics of that portion of the dataset, and a  \texttt{LICENCE.txt} that specifies the exact usage rights for the given portion of the dataset (this depends on the originating institution, so it is easiest to apply to all its pages in bulk).
\subsection{Metadata}
For each of the pages included in the dataset, characterising information was collected for the sake of enabling fine-grained evaluation of models --- practitioners can pick subsets of data that follow specific characteristics they might be interested in ---, musicological contextualisation of models' behaviour and potential future extensions to the use cases this dataset can support. This information is provided on the page-level folder within a \texttt{metadata.json} file.

A first block of metadata is dedicated to the identification of the musical document. Although some field names have been adapted to suit the needs of this project, several elements in the metadata file were defined in accordance with the cataloguing standards of the Répertoire International des Sources Musicales (RISM). This applies to "author" (corresponding to "Heading-Personal name" (100 \$a) in RISM), "title\_description" (matching "Title on source" (245 \$a) in RISM), "institution\_rism\_siglum" (equivalent to "Library siglum" (852 \$a) in RISM), "rism\_id\_number" (matching "RISM ID number", 001), and "shelfmark" (aligned with "Shelfmark" (852 \$c) in RISM). We provide information about the institution that is guarding the manuscript, about the manuscript itself, a link to a digitised version if available and some details about the hands that produced it. The specific fields that have been collected for this block are:
\begin{multicols}{2}
    \begin{itemize}[noitemsep,leftmargin=*]
        \item Full institution name.
        \item Institution RISM siglum.
        \item Manuscript local signature.
        \item Manuscript RISM ID number.
        \item Source date.
        \item Scribal hand.
        \columnbreak
        \item Page size.
        \item Title description.
        \item Author.
        \item Author lifespan.
        \item Genre or form (motet, quartet ...).
    \end{itemize}
\end{multicols}
\noindent Given that the annotation process is page-level, we also include the path to the specific page and the page number within the work as part of this metadata.

A second block is devoted to the notational properties of the musical documents in the collection, offering a broad musicological description of each source, including elements such as notation type and part composition. This information was manually extracted by expert musicologists. The specific fields that have been collected for this block are:
\begin{itemize}[noitemsep,leftmargin=*]
    \item Notation style (CWMN, mensural, square, adiastematic, instrument-specific, other).
    \item Detailed description of the notation, if applicable.
    \item Complexity (\texttt{monophonic}, \texttt{homophonic}, \texttt{polyphonic}, \texttt{pianoform}) \cite{calvo-zaragozaUnderstandingOpticalMusic2021}.
    \item Production (printed, manuscript, born-digital).
    \item Clarity, as in the quality/legibility \cite{alaei_quality_2018} of the document (\texttt{perfect}, \texttt{sufficient}, \texttt{problematic}, \texttt{unreadable}).
    \item Number of systems and staves per system.
\end{itemize}
Some documents were found to be heterogeneous with respect to some of these categories. In this case, the most complex case is reported. For instance, if a page contains a particella with monophonic instruments and a pianoform part, the complexity of the page is set to \texttt{pianoform}.

The definition of the clarity field in this metadata can be a bit ambiguous. The broad guideline followed to define this field is the following. A score with \texttt{perfect} clarity is one that does not have any visible artifacts on the page and the notation is clear and readable. A score with \texttt{sufficient} clarity is one that contains some artifacts but is still easily readable. A score with \texttt{problematic} clarity is one that contains a significant amount of artifacts that make it difficult to read the score, but its contents can still be retrieved fully without ambiguity. Finally, a score with \texttt{unreadable} clarity is one that cannot be unambiguously read and full transcriptions require making assumptions about the piece.


%
\section{Data Overview}
The MusiCorpus v1.0 dataset contains a total of 1,309 pages divided into a train-validation-test split, summarised in \autoref{table:summary}. 
It spans the mid-18th to mid-20th century and contains music from both secular and sacred contexts, with a mix of monophonic, polyphonic, and pianoform notations; predominantly in manuscript form.
The overview of selected works, their provenances, and the features of the music notation can be found in the metadata records for each page in the dataset.

%
\begin{table}[ht]
\caption{Summary of basic stats for the dataset. We log the number of Pages, Systems, Annotated Symbols and Transcribed Notes, these extracted from the COCO and MusicXML files respectively. We also compute the average number of Symbols and Notes both per Page and System for context. This is provided separately for each split and subset and also jointly in both dimensions.}
\label{table:summary}
\resizebox{\linewidth}{!}{
\begin{tabular}{@{}llllllllllllllll@{}}
\toprule
Split                   & \multicolumn{3}{c}{\textbf{Train}}                            & \hspace{12pt} & \multicolumn{3}{c}{\textbf{Validation}}                       & \hspace{12pt} & \multicolumn{3}{c}{\textbf{Test}}                             & \hspace{12pt} & \multicolumn{3}{c}{\textbf{Global}\vspace{6pt}}                           \\
Block                 & \rotatebox{45}{\textit{Dolores}} & \rotatebox{45}{\textit{OmniOMR}} & \rotatebox{45}{\textit{Joint}} & \rotatebox{45}{\textit{}}                     & \rotatebox{45}{\textit{Dolores}} & \rotatebox{45}{\textit{OmniOMR}} & \rotatebox{45}{\textit{Joint}} & \rotatebox{45}{\textit{}}                     & \rotatebox{45}{\textit{Dolores}} & \rotatebox{45}{\textit{OmniOMR}} & \rotatebox{45}{\textit{Joint}} & \rotatebox{45}{\textit{}}                     & \rotatebox{45}{\textit{Dolores}} & \rotatebox{45}{\textit{OmniOMR}} & \rotatebox{45}{\textit{Joint}} \\ \midrule
\textbf{\# Pages}                & 724                  & 60                    & 784            &                               & 186                  & 20                    & 206            &                               & 299                  & 20                    & 319            &                               & 1,209                & 100                   & 1,309          \\
\textbf{\# Systems}              & 5,409                & 264                   & 5,673          &                               & 1,439                & 85                    & 1,524          &                               & 2,195                & 74                    & 2,269          &                               & 9,043                & 423                   & 9,466          \\
\textbf{\# Symbols }             & 465,724              & 65,696                & 531,420        &                               & 119,279              & 18,777                & 138,056        &                               & 191,407              & 20,509                & 211,916        &                               & 776,410              & 104,982               & 881,392        \\
\textbf{Avg. Sym. Page}   & 643.27               & 1094.93               & 677.83         &                               & 641.28               & 938.85                & 670.17         &                               & 640.16               & 1025.45               & 664.31         &                               & 642.19               & 1,049.82               & 673.33         \\
\textbf{Avg. Sym. System} & 86.10                & 248.85                & 93.68          &                               & 82.89                & 220.91                & 90.59          &                               & 87.20                & 277.15                & 93.40          &                               & 85.86                & 248.18                & 93.11          \\
\textbf{\# Notes}                & 207,633              & 19,290                & 226,923        &                               & 53,477               & 5,592                 & 59,069         &                               & 87,066               & 6,338                 & 93,404         &                               & 348,176              & 31,220                & 379,396        \\
\textbf{Avg. Notes Page}     & 286.79               & 321.50                & 289.44         &                               & 287.51               & 279.60                & 286.74         &                               & 291.19               & 316.90                & 292.80         &                               & 287.99               & 312.20                & 289.84         \\
\textbf{Avg. Notes System}   & 38.39                & 73.07                 & 40.00          &                               & 37.16                & 65.79                 & 38.76          &                               & 39.67                & 85.65                 & 41.17          &                               & 38.50                & 73.81                 & 40.08          \\ \bottomrule
\end{tabular}
}
\end{table}

\section{Technical validation}
The data has been validated to ensure correctness, consistency, and utility on several levels. 
Quality control of both adherence to annotation formats and of factual correctness was first performed during the annotation processes and automated postprocessing that generated the dataset structure (see Methods).
Beyond quality control during the dataset creation process, we checked that the finalised datasets in fact allow for training OMR models. We used two baselines: an object detection model, and an image-to-sequence model.

\subsection{Object detection baseline}

Using the YOLOv12 object detection model~\cite{tianYOLOv12AttentionCentricRealTime2025}, we performed object detection on both of the proposed dataset blocks. We train with image patches of $640 \times 640$ pixels.
The results for the main object classes (those that participate in encoding musical semantics: pitches, durations, and onsets of notes), as defined by~\cite{hajicTowardsFullPipeline2018} are reported in \autoref{table:detection-results}.

\begin{table}[ht]
    \centering

    \caption{Object detection results for the main object classes of the OmniOMR and Dolores portions of the dataset on the main symbol classes that participate in visually encoding musical semantics: pitches, durations, and onsets. Counts are given on test split. Aggregate classes in \textit{italic}: ``Barlines'' are a weighted average over SMuFL classes barlineFinal, barlineHeavy, barlineSingle, barlineWing (the overwhelming majority are barlineSingle), with mAP 0.91; ``Flags'' are a weighted average over SMuFL classes flag8thUp, flag8thDown, flag16thUp, flag16thDown, flag32ndUp, and flag32ndDown.
    ``Rests'' is similarly a weighted average over all rest durations (rest16th, rest8th, restQuarter, etc., up to restDoubleWhole and restLonga). Classes with asterisk have no SMuFL codepoint assigned.}
    \label{table:detection-results}

\begin{tabular}{@{}p{0.28\linewidth}p{0.09\linewidth}p{0.12\linewidth}p{0.05\linewidth}p{0.09\linewidth}p{0.12\linewidth}@{}}
\toprule
\multicolumn{1}{c}{\textbf{SMuFL}} & \multicolumn{2}{c}{\textbf{OmniOMR}} &  & \multicolumn{2}{c}{\textbf{Dolores}} \\  
\textbf{}                          & \textit{\#}    & \textit{mAP@50}    &  & \textit{\#}    & \textit{mAP@50}    \\ \midrule
accidentalFlat                     & 15             & 0.98                &  & 1,094          & 0.91                    \\
accidentalNatural                  & 35             & 0.97                &  & 1,577          & 0.94                    \\
accidentalSharp                    & 90             & 0.99                &  & 1,895          & 0.91                    \\
augmentationDot                    & 122            & 0.95                &  & 4,768          & 0.72                    \\
\textit{barlines}                           & 598            & 0.90                &  & 21,458         & 0.78                    \\
beam*                               & 481            & 0.96                &  & 12,457         & 0.88                    \\
cClef & 11 & 0.88 & & 86 & 0.48 \\
fClef                              & 10             & 0.92                &  & 81            & 0.73                    \\
\textit{flags}                              & 198            & 0.87                &  & 6,453          & 0.87                    \\
gClef                              & 42             & 0.87                &  & 141           & 0.73                    \\
noteheadBlack                      & 1,794           & 0.99                &  & 52,755         & 0.87                    \\
noteheadHalf                       & 95             & 0.95                &  & 7,275          & 0.93                    \\
noteheadWhole                      & 18             & 0.77                &  & 1,591          & 0.95                    \\
\textit{rests}                              & 249            & 0.88                &  & 13,160         & 0.87                    \\
stem                               & 1,734           & 0.93                &  & 56,232         & 0.78                    \\ \bottomrule
\end{tabular}
\end{table}

The results closely follow the expected distribution of results from handwritten sheet music object detection \cite{hajicTowardsFullPipeline2018,pachaBaselineGeneralMusicObject2018}: high scores for frequent classes with very visually consistent glyphs (noteheads, leger lines, augmentation dot), lower scores for less frequent, more visually complex classes such as clefs.

\subsection{Sequence-to-sequence baseline}

To validate the quality of the transcriptions on the dataset we have implemented a simple baseline inspired by the way Handwritten Music Recognition of CWMN has been performed in the literature \cite{baroOpticalMusicRecognition2019,baroHandwrittenHistoricalMusic2020,torrasIntegrationLanguageModels2021a}. In particular, we have trained a line-level HTR-VT model \cite{liHTRVTHandwrittenText2025a} on the notation format seen in \cite{baroOpticalMusicRecognition2019} (which in its turn was spawned from \cite{calvo-zaragozaCameraPrIMuSNeuralEndtoEnd2018}) and computed a simple Symbol Error Rate (SER) metric to have an assessment of the quality of transcription that can be attained using this dataset, defined as
\begin{equation}
    SER(\hat{y}, y) = \frac{I + S + D}{|y|}
\end{equation}
where $\hat{y}$ and $y$ are the model's prediction and the ground truth, respectively and I, S and D are the number of insertions, substitutions and deletions required to get the ground truth sequence of symbols from the predicted sequence of symbols.

HTR-VT is an evolution of the classic Recurrent Neural Network (RNN) with Connectionist Temporal Classification (CTC) recognition architecture, incorporating advances such as replacing RNNs with Transformer layers \cite{vaswaniAttentionAllYou2017} or employing Masked Autoencoding in its training procedure. Its main strength is the fact that it can reach very high recognition results at a relatively low computational cost --- both in terms of training data requirements and raw model size --- while staying compatible with existing HMR pipelines. To train this model we have employed the subset of monophonic, line-level samples on the dataset, converted them to the same format as \cite{baroHandwrittenHistoricalMusic2020} and fed them to the model. The number of usable samples for this experiment is around 6,313 from a pool of 7,074 training staff-level lines.

The model uses most of the default hyperparameters from HTR-VT~\cite{liHTRVTHandwrittenText2025a}. It is based on a ViT backbone of 4 layers of dimension 768 with 6 different attention heads and a feed-forward layer dimension of 3072. The HTR-VT masking probability ratio is kept to the default 0.4 with a span length of 8. We adjust the model's input image size to be $1024 \times 64$ to accommodate the aspect ratio of the line-level music samples, but we keep the same data augmentations. The model is trained using AdamW for 100,000 iterations with a weight decay of 0.4 and a warm-up-cosine learning rate schedule with a max learning rate of $1e-3$ and 1,000 warm-up iterations.

The final results on test yielded a 27.14\% of SER on the monophonic, single-staff samples, which are 2,399 out of the total available 2,664 staff-level samples.

\section{Usage Notes}

The dataset is available at the following stable URL: \url{http://hdl.handle.net/20.500.12800/1-6147}.\footnote{LINDAT/CLARIN-EH is a repository of scientific data and tools for the digital humanities and language data, identifying items using the Handle system.}

\section{Acknowledgements}

This work has been partially supported by the Spanish projects
CNS2022-135947 (DOLORES), PID2021-126808OB-I00 (GRAIL) and PID2024-157778OB-I00 \\ (SUKIDI) from the Ministerio de Ciencia e Innovación, the grant Càtedra ENIA UAB-Cruïlla (TSI-100929-2023-2) from the Ministry of Economic Affairs and Digital Transition of Spain, the Departament de Cultura of the Generalitat de Catalunya, and the CERCA Program / Generalitat de Catalunya. Pau Torras is funded by the Spanish
FPU Grant FPU22/00207. Alicia Fornés acknowledges the financial support for her general research activities from ICREA under the ICREA Academia (Departament de Recerca i Universitats de la Generalitat de Catalunya). 
We gratefully thank Josep Tardío, Edgar Zamora, Irene Andreu, Ana Valeria Blanco, Àngela Martin, Pedro Jesús López, Carlota Franch, Laura Marbà, Anna Fusté, Raquel Garcia, Laura Andrea Monroy and Jialuo Chen for their assistance in this work.

This work has been further supported by the Ministry of Culture of the Czech Republic (project OmniOMR of the NAKI III programme, no. DH23P03OVV008), by the project “Human-centred AI for a Sustainable and Adaptive Society” (reg. no. CZ.02.01.01/00/23\_025/0008691), co-funded by the European Union, by the Charles University (project GAUK no. 289623), by SVV project no. 260 821, and by the Ministry of Education, Youth and Sports of the Czech Republic, project no. LM2023062 LINDAT/CLARIAH-CZ. 
We wish to thank Hana Bečková, Eva Kolářová, Anna Cibulková, Kateřina Maňáková, Tereza Svatošová, Michael Pinkas, Daria Savateeva, Jan Kravárik, Šárka Mayerová, and Čeněk Řehoř for their assistance in this work.

We gratefully acknowledge the involvement of the institutions that have granted access to their material and allowed sharing it publicly in order to build this dataset. We thank the Xarxa d'Arxius Comarcals de Catalunya, the Societat del Gran Teatre del Liceu, the Institució Milà i Fontanals of the Spanish National Research Council, the Documentation Centre of the Palau de la Música Catalana and the Moravian Library.

\section{Author Contributions}

\textbf{Pau Torras:} Conceptualization, Data curation, Formal analysis, Investigation, Methodology, Project administration, Software, Validation, Visualization, Writing – original draft, Writing – review \& editing.

\noindent
\textbf{Ji\v{r}\'{i} Mayer:} Conceptualization, Data curation, Formal analysis, Investigation, Methodology, Project administration, Software, Validation, Visualization, Writing – original draft, Writing – review \& editing.

\noindent
\textbf{Carles Badal:} Data curation, Investigation, Methodology, Project administration, Resources, Validation, Writing – review \& editing.

\noindent
\textbf{Martina Dvo\v{r}\'{a}kov\'{a}:} Data curation, Investigation, Methodology, Project administration, Resources, Validation, Writing – review \& editing.

\noindent
\textbf{Markéta Herzánová Vlková:} Data curation, Investigation, Methodology, Project administration, Resources, Validation, Writing – review \& editing.

\noindent
\textbf{Gerard Asbert:} Data curation, Investigation, Software, Writing – review \& editing.

\noindent
\textbf{Vojtěch Dvořák:} Formal analysis, Investigation, Software, Validation, Visualization, Writing – review \& editing.

\noindent
\textbf{Samuel Šomorjai:} Data curation, Investigation, Project administration, Resources, Validation, Writing – review \& editing.

\noindent
\textbf{Jan Hajič jr.:} Conceptualization, Funding acquisition, Investigation, Methodology, Project administration, Resources, Supervision, Visualization, Writing – original draft, Writing – review \& editing.

\noindent
\textbf{Alicia Fornés:} Conceptualization, Funding acquisition, Investigation, Methodology, Project administration, Resources, Supervision, Writing – original draft, Writing – review \& editing.


\bibliography{bibliography}

@inproceedings{baroHandwrittenHistoricalMusic2020,
  title = {Handwritten {{Historical Music Recognition}} by {{Sequence-to-Sequence}} with {{Attention Mechanism}}},
  booktitle = {2020 17th {{International Conference}} on {{Frontiers}} in {{Handwriting Recognition}} ({{ICFHR}})},
  author = {Bar{\'o}, Arnau and Badal, Carles and Forn{\^e}s, Alicia},
  year = 2020,
  month = sep,
  pages = {205--210},
  publisher = {IEEE Computer Society},
  address = {Dortmund, Germany},
  doi = {10.1109/ICFHR2020.2020.00046}
}

@inproceedings{calvo-zaragozaCameraPrIMuSNeuralEndtoEnd2018,
  title = {Camera-{{PrIMuS}}: {{Neural End-to-End Optical Music Recognition}} on {{Realistic Monophonic Scores}}},
  booktitle = {18th {{International Society}} for {{Music Information Retrieval Conference}}},
  author = {{Calvo-Zaragoza}, Jorge and Rizo, David},
  year = 2018,
  pages = {248--255},
  publisher = {International Society for Music Information Retrieval},
  address = {Paris, France},
  isbn = {978-2-9540351-2-3}
}

@article{calvo-zaragozaEndtoEndNeuralOptical2018,
  title = {End-to-{{End Neural Optical Music Recognition}} of {{Monophonic Scores}}},
  author = {{Calvo-Zaragoza}, Jorge and Rizo, David},
  year = 2018,
  month = apr,
  journal = {Applied Sciences},
  volume = {8},
  number = {4},
  pages = {606},
  publisher = {Multidisciplinary Digital Publishing Institute},
  issn = {2076-3417},
  doi = {10.3390/app8040606},
  urldate = {2022-10-25},
  copyright = {http://creativecommons.org/licenses/by/3.0/},
  langid = {english}
}

@article{calvo-zaragozaUnderstandingOpticalMusic2021,
  title = {Understanding {{Optical Music Recognition}}},
  author = {{Calvo-Zaragoza}, Jorge and Haji{\v c} Jr., Jan and Pacha, Alexander},
  year = 2021,
  month = jul,
  journal = {ACM Comput. Surv.},
  volume = {53},
  number = {4},
  pages = {1--35},
  issn = {0360-0300, 1557-7341},
  doi = {10.1145/3397499},
  urldate = {2022-10-26},
  langid = {english}
}

@article{fornesCVCMUSCIMAGroundTruth2012,
  title = {{{CVC-MUSCIMA}}: A Ground Truth of Handwritten Music Score Images for Writer Identification and Staff Removal},
  shorttitle = {{{CVC-MUSCIMA}}},
  author = {Forn{\'e}s, Alicia and Dutta, Anjan and Gordo, Albert and Llad{\'o}s, Josep},
  year = 2012,
  month = sep,
  journal = {IJDAR},
  volume = {15},
  number = {3},
  pages = {243--251},
  issn = {1433-2833, 1433-2825},
  doi = {10.1007/s10032-011-0168-2},
  urldate = {2022-10-25},
  langid = {english}
}

@article{fuentes-martinezAlignedMusicNotation2026,
  title = {Aligned Music Notation and Lyrics Transcription},
  author = {{Fuentes-Mart{\'i}nez}, Eliseo and {R{\'i}os-Vila}, Antonio and {Martinez-Sevilla}, Juan C. and Rizo, David and {Calvo-Zaragoza}, Jorge},
  year = 2026,
  month = feb,
  journal = {Pattern Recognition},
  volume = {170},
  pages = {112094},
  issn = {0031-3203},
  doi = {10.1016/j.patcog.2025.112094},
  urldate = {2026-01-19}
}

@inproceedings{hajicMUSCIMADatasetHandwritten2017,
  title = {The {{MUSCIMA}}++ {{Dataset}} for {{Handwritten Optical Music Recognition}}},
  booktitle = {2017 14th {{IAPR International Conference}} on {{Document Analysis}} and {{Recognition}} ({{ICDAR}})},
  author = {Haji{\v c}, Jan and Pecina, Pavel},
  year = 2017,
  month = nov,
  volume = {01},
  pages = {39--46},
  issn = {2379-2140},
  doi = {10.1109/ICDAR.2017.16}
}

@misc{martinez-sevillaSheetMusicBenchmark2025b,
  title = {Sheet {{Music Benchmark}}: {{Standardized Optical Music Recognition Evaluation}}},
  shorttitle = {Sheet {{Music Benchmark}}},
  author = {{Martinez-Sevilla}, Juan C. and {Cerveto-Serrano}, Joan and Luna, Noelia and Chapman, Greg and Sapp, Craig and Rizo, David and {Calvo-Zaragoza}, Jorge},
  year = 2025,
  month = jun,
  number = {arXiv:2506.10488},
  eprint = {2506.10488},
  primaryclass = {cs},
  publisher = {arXiv},
  doi = {10.48550/arXiv.2506.10488},
  urldate = {2026-01-19},
  archiveprefix = {arXiv}
}

@inproceedings{mayerMuNGStudioAnnotation2025,
  title = {{{MuNG Studio}}: {{Annotation Tool}} for {{Music Notation Graph}}},
  shorttitle = {{{MuNG Studio}}},
  booktitle = {Proceedings of the 12th {{International Conference}} on {{Digital Libraries}} for {{Musicology}}},
  author = {Mayer, Ji{\v r}{\'i} and Jebav{\'y}, Filip and Herz{\'a}nov{\'a} Vlkov{\'a}, Mark{\'e}ta and Dvo{\v r}{\'a}kov{\'a}, Martina and Pecina, Pavel and Haji{\v c}, Jan},
  year = 2025,
  month = sep,
  series = {{{DLfM}} '25},
  pages = {114--118},
  publisher = {Association for Computing Machinery},
  address = {New York, NY, USA},
  doi = {10.1145/3748336.3748379},
  urldate = {2025-09-29},
  isbn = {979-8-4007-2083-3}
}

@inproceedings{mayerSmashcimaFullPageHandwritten2025,
  title = {Smashcima: {{Full-Page Handwritten Music Document Synthesizer}}},
  shorttitle = {Smashcima},
  booktitle = {Proceedings of the 12th {{International Conference}} on {{Digital Libraries}} for {{Musicology}}},
  author = {Mayer, Ji{\v r}{\'i} and Pecina, Pavel and Haji{\v c}, Jan},
  year = 2025,
  month = sep,
  series = {{{DLfM}} '25},
  pages = {119--123},
  publisher = {Association for Computing Machinery},
  address = {New York, NY, USA},
  doi = {10.1145/3748336.3748380},
  urldate = {2025-09-29},
  isbn = {979-8-4007-2083-3}
}

@inproceedings{parada-cabaleiroSEILSDatasetSymbolically2017,
  title = {The {{SEILS Dataset}}: {{Symbolically Encoded Scores}} in {{Modern-Early Notation}} for {{Computational Musicology}}},
  booktitle = {18th {{International Society}} for {{Music Information Retrieval Conference}}},
  author = {{Parada-Cabaleiro}, Emilia and Batliner, Anton and Baird, Alice and Schuller, Bj{\"o}rn},
  year = 2017,
  address = {Suzhou, China},
  isbn = {978-981-11-5179-8}
}

@article{rios-vilaEndtoEndFullPageOptical2026,
  title = {End-to-{{End Full-Page Optical Music Recognition}} for {{Pianoform Sheet Music}}},
  author = {{R{\'i}os-Vila}, Antonio and {Calvo-Zaragoza}, Jorge and Rizo, David and Paquet, Thierry},
  year = 2026,
  month = jan,
  journal = {Int J Comput Vis},
  volume = {134},
  number = {2},
  pages = {49},
  issn = {1573-1405},
  doi = {10.1007/s11263-025-02654-6},
  urldate = {2026-01-19},
  langid = {english}
}

@inproceedings{shatriDoReMiFirstGlance2021,
  title = {{{DoReMi}}: {{First}} Glance at a Universal {{OMR}} Dataset},
  booktitle = {Proceedings of the 3rd {{International Workshop}} on {{Reading Music Systems}}},
  author = {Shatri, Elona and Fazekas, Gy{\"o}rgy},
  editor = {{Calvo-Zaragoza}, Jorge and Pacha, Alexander},
  year = 2021,
  pages = {43--49},
  publisher = {International Society for Music Information Retrieval},
  address = {Alicante, Spain}
}

@article{torrasUnifiedRepresentationFramework2024,
  title = {A {{Unified Representation Framework}} for the {{Evaluation}} of {{Optical Music Recognition Systems}}},
  author = {Torras, Pau and Biswas, Sanket and Forn{\'e}s, Alicia},
  year = 2024,
  month = jul,
  journal = {IJDAR},
  volume = {27},
  number = {Special issue on ``advanced topics in document analysis and recognition''},
  pages = {379--393},
  issn = {1433-2825},
  doi = {10.1007/s10032-024-00485-8},
  urldate = {2024-07-24},
  langid = {english}
}

@inproceedings{tuggenerDeepScoresDatasetSegmentation2018,
  title = {{{DeepScores}} - {{A Dataset}} for {{Segmentation}}, {{Detection}} and {{Classification}} of {{Tiny Objects}}},
  booktitle = {24th {{International Conference}} on {{Pattern Recognition}}},
  author = {Tuggener, Lukas and Elezi, Ismail and Schmidhuber, J{\"u}rgen and Pelillo, Marcello and Stadelmann, Thilo},
  year = 2018,
  pages = {3704--3709},
  publisher = {IEEE Computer Society},
  address = {Beijing, China},
  doi = {10.21256/zhaw-4255}
}

@inproceedings{tuggenerDeepScoresV2DatasetBenchmark2020,
  title = {The {{DeepScoresV2 Dataset}} and {{Benchmark}} for {{Music Object Detection}}},
  booktitle = {Proceedings of the 25th {{International Conference}} on {{Pattern Recognition}}},
  author = {Tuggener, Lukas and Satyawan, Yvan Putra and Pacha, Alexander and Schmidhuber, J{\"u}rgen and Stadelmann, Thilo},
  year = 2020,
  pages = {9188--9195},
  publisher = {IEEE Computer Society},
  address = {Milan, Italy},
  doi = {10.21256/zhaw-20647}
}

@inproceedings{knopkeMusicdiffFoundationImproved2007,
    address = {Vienna, Austria},
    title = {Towards {Musicdiff} : {A} {Foundation} for {Improved} {Optical} {Music} {Recognition} {Using} {Multiple} {Recognizers}},
    isbn = {978-3-85403-218},
    url = {http://homes.sice.indiana.edu/donbyrd/Papers/ismir_2007_omr.pdf},
    booktitle = {8th {International} {Conference} on {Music} {Information} {Retrieval}},
    author = {Knopke, Ian and Byrd, Donald},
    year = {2007},
    pages = {123--126},
}

@article{byrdStandardTestbedOptical2015,
    title = {Towards a {Standard} {Testbed} for {Optical} {Music} {Recognition}: {Definitions}, {Metrics}, and {Page} {Images}},
    volume = {44},
    issn = {0929-8215, 1744-5027},
    shorttitle = {Towards a {Standard} {Testbed} for {Optical} {Music} {Recognition}},
    url = {http://www.tandfonline.com/doi/full/10.1080/09298215.2015.1045424},
    doi = {10.1080/09298215.2015.1045424},
    language = {en},
    number = {3},
    urldate = {2022-10-25},
    journal = {Journal of New Music Research},
    author = {Byrd, Donald and Simonsen, Jakob Grue},
    month = jul,
    year = {2015},
    pages = {169--195},
}

@article{rios-vilaEndtoendOpticalMusic2023,
    title = {End-to-end optical music recognition for pianoform sheet music},
    volume = {26},
    issn = {1433-2825},
    url = {https://doi.org/10.1007/s10032-023-00432-z},
    doi = {10.1007/s10032-023-00432-z},
    language = {en},
    number = {3},
    urldate = {2023-09-27},
    journal = {International Journal on Document Analysis and Recognition (IJDAR)},
    author = {Ríos-Vila, Antonio and Rizo, David and Iñesta, José M. and Calvo-Zaragoza, Jorge},
    month = sep,
    year = {2023},
    pages = {347--362},
}

@inproceedings{TIEDEMANN12.463,
    address = {Istanbul, Turkey},
    title = {Parallel data, tools and interfaces in {OPUS}},
    isbn = {978-2-9517408-7-7},
    booktitle = {Proceedings of the eight international conference on language resources and evaluation ({LREC}'12)},
    publisher = {European Language Resources Association (ELRA)},
    author = {Tiedemann, Jörg},
    year = {2012},
}

@unpublished{rigaux:hal-05515751,
    title = {The {CollabScore} dataset -towards robust and generalized {OMR} evaluation},
    url = {https://hal.science/hal-05515751},
    doi = {10.1145/nnnnnnn.nnnnnnn},
    author = {Rigaux, Philippe and Coüasnon, Bertrand and Guillotel-Nothmann, Christophe and Guilloux, Fabien and Lemaitre, Aurélie},
    month = jan,
    year = {2026},
    note = {tex.hal\_id: hal-05515751
tex.hal\_version: v1},
}

@article{Dorfer2018MSMD,
  title = {Learning Audio–Sheet Music Correspondences for Cross-Modal Retrieval and Piece Identification},
  volume = {1},
  ISSN = {2514-3298},
  url = {http://dx.doi.org/10.5334/tismir.12},
  DOI = {10.5334/tismir.12},
  number = {1},
  journal = {Transactions of the International Society for Music Information Retrieval},
  publisher = {Ubiquity Press,  Ltd.},
  author = {Dorfer,  Matthias and Hajič jr.,  Jan and Arzt,  Andreas and Frostel,  Harald and Widmer,  Gerhard},
  year = {2018},
  month = sep,
  pages = {22}
}

@inproceedings{calvo-zaragozaRecognitionPenBasedMusic2014,
    address = {Stockholm, Sweden},
    title = {Recognition of {Pen}-{Based} {Music} {Notation}: {The} {HOMUS} {Dataset}},
    issn = {1051-4651},
    doi = {10.1109/ICPR.2014.524},
    booktitle = {22nd {International} {Conference} on {Pattern} {Recognition}},
    publisher = {IEEE Computer Society},
    author = {Calvo-Zaragoza, Jorge and Oncina, Jose},
    year = {2014},
    pages = {3038--3043},
}

@inproceedings{martinez-sevillaPerformanceOpticalMusic2023,
    title = {On {The} {Performance} of {Optical} {Music} {Recognition} in the {Absence} of {Specific} {Training} {Data}},
    language = {en},
    author = {Martinez-Sevilla, Juan C and Rosello, Adrian and Rizo, David and Calvo-Zaragoza, Jorge},
    year = {2023},
    publisher    = {ISMIR},
    booktitle    = {Proceedings of the 24th International Society for
                    Music Information Retrieval Conference
                   },
    pages = {319--326},
}

@inproceedings{martinez-sevillaTowardsUniversalOptical2024,
  author       = {Juan Carlos Martinez-Sevilla and
                  David Rizo and
                  Jorge Calvo-Zaragoza},
  title        = {Towards Universal Optical Music Recognition: A
                   Case Study on Notation Types
                  },
  booktitle    = {Proceedings of the 25th International Society for
                   Music Information Retrieval Conference
                  },
  year         = 2024,
  pages        = {914-921},
  publisher    = {ISMIR},
  month        = nov,
  venue        = {San Francisco, California, USA and Online},
  doi          = {10.5281/zenodo.14877479},
  url          = {https://doi.org/10.5281/zenodo.14877479},
}

@inproceedings{calvo-zaragozaTwoNoteHeads2016,
    address = {New York City},
    title = {Two (note) heads are better than one: pen-based multimodal interaction with music scores},
    isbn = {978-0-692-75506-8},
    url = {https://wp.nyu.edu/ismir2016/wp-content/uploads/sites/2294/2016/07/006_Paper.pdf},
    booktitle = {17th {International} {Society} for {Music} {Information} {Retrieval} {Conference}},
    publisher = {International Society for Music Information Retrieval},
    author = {Calvo-Zaragoza, Jorge and Rizo, David and Iñesta, José Manuel},
    editor = {Devaney, J. et al.},
    year = {2016},
    pages = {509--514},
}

@inproceedings{mayerPracticalEndToEnd2023,
    author={Jiří Mayer and Milan Straka and Jan {Hajič jr.} and Pavel Pecina},
    title={Practical End-to-End Optical Music Recognition for Pianoform Music},
    booktitle={18th International Conference on Document Analysis and Recognition ({ICDAR})},
    pages={55-73},
    address={Athens, Greece},
    year={2024},
    doi={10.1007/978-3-031-70552-6_4}
}

@inproceedings{gothamScoresOfScore2018,
    author = {Gotham, Mark and Jonas, Peter and Bower, Bruno and Bosworth, William and Rootham, Daniel and VanHandel, Leigh},
    title = {Scores of scores: an openscore project to encode and share sheet music},
    booktitle = {5th International Conference on Digital Libraries for Musicology (DLfM)},
    pages = {87–95},
    address = {Paris, France},
    year = {2018},
    doi = {10.1145/3273024.3273026},
    
    _isbn = {9781450365222},
    _publisher = {Association for Computing Machinery},
    _numpages = {9},
}

@inproceedings{gothamOpenScoreLieder2022,
    author = {Gotham, Mark Robert Haigh and Jonas, Peter},
    title = {{The OpenScore Lieder Corpus}},
    booktitle = {{Music Encoding Conference}},
    pages = {131--136},
    address = {Alicante, Spain},
    year = {2022},
    doi = {10.17613/1my2-dm23},
    
    _publisher = {{Humanities Commons}},
    _isbn = {978-84-1302-173-7},
    _editor = {M{\"u}nnich, Stefan and Rizo, David},
}

@inproceedings{repoluskKuiSCIMADatasetOptical2024,
    address = {Cham},
    title = {The {KuiSCIMA} {Dataset} for {Optical} {Music} {Recognition} of {Ancient} {Chinese} {Suzipu} {Notation}},
    isbn = {978-3-031-70552-6},
    doi = {10.1007/978-3-031-70552-6_3},
    language = {en},
    booktitle = {Document {Analysis} and {Recognition} - {ICDAR} 2024},
    publisher = {Springer Nature Switzerland},
    author = {Repolusk, Tristan and Veas, Eduardo},
    editor = {Barney Smith, Elisa H. and Liwicki, Marcus and Peng, Liangrui},
    year = {2024},
    pages = {38--54},
}

@inproceedings{hajicFurtherSteps2016,
    author = {Haji{\v{c}} jr., Jan and Novotn\'{y}, Ji\v{r}\'{i} and Pecina, Pavel and Pokorn\'{y}, Jaroslav},
    title = {Further Steps towards a Standard Testbed for Optical Music Recognition},
    booktitle = {17th International Society for Music Information Retrieval Conference (ISMIR)},
    pages = {157--163},
    address = {New York, {USA}},
    year = {2016},
    url = {https://wp.nyu.edu/ismir2016/event/proceedings/},
    
    _editor = {Michael Mandel and Johanna Devaney and Douglas Turnbull and George Tzanetakis},
    _organization = {New York University},
    _publisher = {New York University},
    _isbn = {978-0-692-75506-8},
}

@article{lavinTechnologyReadinessLevelsML2022,
  title = {Technology readiness levels for machine learning systems},
  volume = {13},
  ISSN = {2041-1723},
  url = {http://dx.doi.org/10.1038/s41467-022-33128-9},
  DOI = {10.1038/s41467-022-33128-9},
  number = {1},
  journal = {Nature Communications},
  publisher = {Springer Science and Business Media LLC},
  author = {Lavin,  Alexander and Gilligan-Lee,  Ciarán M. and Visnjic,  Alessya and Ganju,  Siddha and Newman,  Dava and Ganguly,  Sujoy and Lange,  Danny and Baydin,  Atílím G\"{u}neş and Sharma,  Amit and Gibson,  Adam and Zheng,  Stephan and Xing,  Eric P. and Mattmann,  Chris and Parr,  James and Gal,  Yarin},
  year = {2022},
  month = oct 
}

@article{rebeloOMRStateOfTheArt2012,
  title = {{Optical Music Recognition}: state-of-the-art and open issues},
  volume = {1},
  ISSN = {2192-662X},
  url = {http://dx.doi.org/10.1007/s13735-012-0004-6},
  DOI = {10.1007/s13735-012-0004-6},
  number = {3},
  journal = {International Journal of Multimedia Information Retrieval},
  publisher = {Springer Science and Business Media LLC},
  author = {Rebelo,  Ana and Fujinaga,  Ichiro and Paszkiewicz,  Filipe and Marcal,  Andre R. S. and Guedes,  Carlos and Cardoso,  Jaime S.},
  year = {2012},
  month = mar,
  pages = {173–190}
}

@inproceedings{dvorakStaffLayoutAnalysis2024,
    author = {Dvořák, Vojtěch and Hajič jr., Jan and Mayer, Jiří},
    title  = {Staff Layout Analysis Using the {YOLO} Platform},
    booktitle = {6th International Workshop on Reading Music Systems ({WoRMS})},
    pages = {18-22},
    address = {Online},
    year = {2024},
    doi={10.48550/arXiv.2411.15741}
}

@article{gallegoStaffLineRemoval2017,
  title={Staff-line removal with selectional auto-encoders},
  author={Gallego, Antonio-Javier and Calvo-Zaragoza, Jorge},
  journal={Expert Systems with Applications},
  volume={89},
  pages={138--148},
  year={2017},
  publisher={Elsevier}
}

@inproceedings{hajicTowardsFullPipeline2018,
    author = {Haji{\v{c}} jr., Jan and Dorfer, Matthias and Widmer, Gerhard and Pecina, Pavel},
    title = {Towards Full-Pipeline Handwritten {OMR} with Musical Symbol Detection by U-Nets},
    booktitle = {19th International Society for Music Information Retrieval Conference (ISMIR)},
    pages = {225--232},
    address = {Paris, France},
    year = {2018},
    url = {http://ismir2018.ircam.fr/doc/pdfs/175_Paper.pdf},
    
    _isbn = {978-2-9540351-2-3},
}

@article{pachaBaselineGeneralMusicObject2018,
    author = {Alexander Pacha and Jan {Hajič jr.} and Jorge Calvo-Zaragoza},
    title = {A Baseline for General Music Object Detection with Deep Learning},
    journal = {Applied Sciences},
    volume = {8},
    number={9},
    pages = {1488-1508},
    year = {2018},
    doi = {10.3390/app8091488}
}

@article{harteltOpticalMedievalMusicRecognition2024,
  title={Optical {Medieval} {Music} {Recognition} — {A} {Complete} {Pipeline} for {Historic} {Chants}.},
  author={Hartelt, Alexander and Eipert, Tim and Puppe, Frank},
  journal={Applied Sciences (2076-3417)},
  volume={14},
  number={16},
  year={2024}
}

@inproceedings{fujinagaArtOfTeachingComputersSIMSSA2019,
  title={The art of teaching computers: the SIMSSA optical music recognition workflow system},
  author={Fujinaga, Ichiro and Vigliensoni, Gabriel},
  booktitle={2019 27th European Signal Processing Conference (EUSIPCO)},
  pages={1--5},
  year={2019},
  organization={IEEE}
}

@misc{puginVerovio,
    title = {Verovio},
    url = {https://www.verovio.org/},
    language = {en},
    urldate = {2024-03-14},
    journal = {Verovio},
    author = {Pugin, Laurent},
}

@article{baroOpticalMusicRecognition2019,
    title = {From {Optical} {Music} {Recognition} to {Handwritten} {Music} {Recognition}: {A} baseline},
    volume = {123},
    issn = {01678655},
    shorttitle = {From {Optical} {Music} {Recognition} to {Handwritten} {Music} {Recognition}},
    url = {https://linkinghub.elsevier.com/retrieve/pii/S0167865518303386},
    doi = {10.1016/j.patrec.2019.02.029},
    language = {en},
    urldate = {2022-11-14},
    journal = {Pattern Recognition Letters},
    author = {Baró, Arnau and Riba, Pau and Calvo-Zaragoza, Jorge and Fornés, Alicia},
    month = may,
    year = {2019},
    pages = {1--8},
}

@inproceedings{alaei_quality_2018,
    address = {Vienna},
    title = {"{Quality}" vs. "{Readability}" in {Document} {Images}: {Statistical} {Analysis} of {Human} {Perception}},
    isbn = {978-1-5386-3346-5},
    shorttitle = {"{Quality}" vs. "{Readability}" in {Document} {Images}},
    url = {https://ieeexplore.ieee.org/document/8395223/},
    doi = {10.1109/DAS.2018.42},
    language = {en},
    urldate = {2026-03-17},
    booktitle = {2018 13th {IAPR} {International} {Workshop} on {Document} {Analysis} {Systems} ({DAS})},
    publisher = {IEEE},
    author = {Alaei, Alireza and Raveaux, Romain and Conte, Donatello and Stantic, Bela},
    month = apr,
    year = {2018},
    pages = {363--368},
}

@inproceedings{torrasIntegrationLanguageModels2021a,
    address = {Online},
    title = {On the {Integration} of {Language} {Models} into {Sequence} to {Sequence} {Architectures} for {Handwritten} {Music} {Recognition}},
    url = {https://zenodo.org/record/5624451},
    doi = {10.5281/zenodo.5624451},
    urldate = {2022-11-15},
    booktitle = {Proceedings of the 22nd {International} {Society} for {Music} {Information} {Retrieval} {Conference}},
    publisher = {ISMIR},
    author = {Torras, Pau and Baró, Arnau and Kang, Lei and Fornés, Alicia},
    month = nov,
    year = {2021},
    pages = {690--696},
}

@misc{fuentes-martinezAlignedMusicNotation2024,
    title = {Aligned {Music} {Notation} and {Lyrics} {Transcription}},
    url = {http://arxiv.org/abs/2412.04217},
    doi = {10.48550/arXiv.2412.04217},
    urldate = {2024-12-08},
    publisher = {arXiv},
    author = {Fuentes-Martínez, Eliseo and Ríos-Vila, Antonio and Martinez-Sevilla, Juan C. and Rizo, David and Calvo-Zaragoza, Jorge},
    month = dec,
    year = {2024},
    note = {arXiv:2412.04217 [cs]},
}

@article{garrido-munozHolisticApproachImagetograph2022,
    title = {A holistic approach for image-to-graph: application to optical music recognition},
    issn = {1433-2825},
    shorttitle = {A holistic approach for image-to-graph},
    url = {https://doi.org/10.1007/s10032-022-00417-4},
    doi = {10.1007/s10032-022-00417-4},
    language = {en},
    urldate = {2022-10-25},
    journal = {International Journal on Document Analysis and Recognition (IJDAR)},
    author = {Garrido-Munoz, Carlos and Rios-Vila, Antonio and Calvo-Zaragoza, Jorge},
    month = sep,
    year = {2022},
}

@inproceedings{rios-vilaEndToEndFullPageOptical2022,
    address = {Online},
    title = {End-{To}-{End} {Full}-{Page} {Optical} {Music} {Recognition} of {Monophonic} {Documents} via {Score} {Unfolding}},
    url = {https://sites.google.com/view/worms2022/proceedings},
    doi = {10.48550/arXiv.2211.13285},
    booktitle = {Proceedings of the 4th {International} {Workshop} on {Reading} {Music} {Systems}},
    author = {Ríos-Vila, Antonio and Iñesta, Jose M. and Calvo-Zaragoza, Jorge},
    editor = {Calvo-Zaragoza, Jorge and Pacha, Alexander and Shatri, Elona},
    year = {2022},
    pages = {20--24},
}

@unpublished{lemaitre_collabscore_2026,
    title = {{CollabScore} {OMR}: {A} {Hybrid} {System} for {Music} {Score} {Recognition}},
    shorttitle = {{CollabScore} {OMR}},
    url = {https://hal.science/hal-05545419},
    urldate = {2026-03-18},
    author = {Lemaitre, Aurélie and Coüasnon, Bertrand and Rigaux, Philippe},
    month = feb,
    year = {2026},
    note = {Submitted to ICDAR 2026},
}

@inproceedings{10.1145/3543882.3543885,
    address = {Prague, Czech Republic},
    series = {{DLfM} '22},
    title = {Digitization of choirbooks in guatemala},
    isbn = {978-1-4503-9668-4},
    url = {https://doi.org/10.1145/3543882.3543885},
    doi = {10.1145/3543882.3543885},
    booktitle = {Proceedings of the 9th international conference on digital libraries for musicology},
    publisher = {Association for Computing Machinery},
    author = {Thomae, Martha E. and Cumming, Julie E. and Fujinaga, Ichiro},
    year = {2022},
    note = {Number of pages: 8
tex.address: New York, NY, USA},
    pages = {19--26},
}

@misc{martinez-sevillaOpticalMusicRecognition2025,
    title = {Optical {Music} {Recognition} of {Jazz} {Lead} {Sheets}},
    url = {http://arxiv.org/abs/2509.05329},
    doi = {10.48550/arXiv.2509.05329},
    urldate = {2026-03-18},
    publisher = {arXiv},
    author = {Martinez-Sevilla, Juan Carlos and Foscarin, Francesco and Garcia-Iasci, Patricia and Rizo, David and Calvo-Zaragoza, Jorge and Widmer, Gerhard},
    month = aug,
    year = {2025},
    note = {arXiv:2509.05329 [cs]},
}

@article{kimAutomaticRecognitionOfJeongganbo2025,
    author = {Kim, Dongmin and Han, Danbinaerin and Jeong, Dasaem and Valero-Mas, Jose J.},
    title = {On the Automatic Recognition of Jeongganbo Music Notation: Dataset and Approach},
    year = {2025},
    issue_date = {September 2025},
    publisher = {Association for Computing Machinery},
    address = {New York, NY, USA},
    volume = {18},
    number = {3},
    issn = {1556-4673},
    url = {https://doi.org/10.1145/3715159},
    doi = {10.1145/3715159},
    journal = {J. Comput. Cult. Herit.},
    month = sep,
    articleno = {52},
    numpages = {21}
}

@inproceedings{pachaLearningNotationGraph2019,
  author       = {Alexander Pacha and
                  Jorge Calvo{-}Zaragoza and
                  Jan Hajic jr.},
  editor       = {Arthur Flexer and
                  Geoffroy Peeters and
                  Juli{\'{a}}n Urbano and
                  Anja Volk},
  title        = {Learning Notation Graph Construction for Full-Pipeline Optical Music
                  Recognition},
  booktitle    = {Proceedings of the 20th International Society for Music Information
                  Retrieval Conference, {ISMIR} 2019, Delft, The Netherlands, November
                  4-8, 2019},
  pages        = {75--82},
  year         = {2019},
  url          = {http://archives.ismir.net/ismir2019/paper/000006.pdf},
  bibsource    = {dblp computer science bibliography, https://dblp.org}
}

@article{couasnonDMOSGenericDocument2006,
    title = {{DMOS}, a generic document recognition method: application to table structure analysis in a general and in a specific way},
    volume = {8},
    issn = {1433-2825},
    shorttitle = {{DMOS}, a generic document recognition method},
    url = {https://doi.org/10.1007/s10032-005-0148-5},
    doi = {10.1007/s10032-005-0148-5},
    language = {en},
    number = {2},
    urldate = {2024-09-17},
    journal = {International Journal of Document Analysis and Recognition (IJDAR)},
    author = {Coüasnon, Bertrand},
    month = jun,
    year = {2006},
    pages = {111--122},
}

@misc{vaswaniAttentionAllYou2017,
    title = {Attention {Is} {All} {You} {Need}},
    url = {http://arxiv.org/abs/1706.03762},
    doi = {10.48550/arXiv.1706.03762},
    urldate = {2022-10-25},
    publisher = {arXiv},
    author = {Vaswani, Ashish and Shazeer, Noam and Parmar, Niki and Uszkoreit, Jakob and Jones, Llion and Gomez, Aidan N. and Kaiser, Lukasz and Polosukhin, Illia},
    month = dec,
    year = {2017},
    note = {arXiv:1706.03762 [cs]
version: 5},
}

@article{liHTRVTHandwrittenText2025a,
    title = {{HTR}-{VT}: {Handwritten} text recognition with vision transformer},
    volume = {158},
    issn = {00313203},
    shorttitle = {{HTR}-{VT}},
    url = {https://linkinghub.elsevier.com/retrieve/pii/S0031320324007180},
    doi = {10.1016/j.patcog.2024.110967},
    language = {en},
    urldate = {2026-03-19},
    journal = {Pattern Recognition},
    author = {Li, Yuting and Chen, Dexiong and Tang, Tinglong and Shen, Xi},
    month = feb,
    year = {2025},
    pages = {110967},
}

@inproceedings{linMicrosoftCOCOCommon2014,
    address = {Cham},
    title = {Microsoft {COCO}: {Common} {Objects} in {Context}},
    isbn = {978-3-319-10602-1},
    url = {https://link.springer.com/chapter/10.1007/978-3-319-10602-1_48},
    booktitle = {Computer {Vision} – {ECCV} 2014},
    publisher = {Springer International Publishing},
    author = {Lin, Tsung-Yi and Maire, Michael and Belongie, Serge and Hays, James and Perona, Pietro and Ramanan, Deva and Dollár, Piotr and Zitnick, C. Lawrence},
    editor = {Fleet, David and Pajdla, Tomas and Schiele, Bernt and Tuytelaars, Tinne},
    year = {2014},
    pages = {740--755},
}

@misc{tianYOLOv12AttentionCentricRealTime2025,
    title = {{YOLOv12}: {Attention}-{Centric} {Real}-{Time} {Object} {Detectors}},
    shorttitle = {{YOLOv12}},
    url = {http://arxiv.org/abs/2502.12524},
    doi = {10.48550/arXiv.2502.12524},
    urldate = {2026-04-29},
    publisher = {arXiv},
    author = {Tian, Yunjie and Ye, Qixiang and Doermann, David},
    month = feb,
    year = {2025},
    note = {arXiv:2502.12524 [cs]},
}
\bibliographystyle{ieeetr}

\newpage

\appendix
\section{Musicorpus data record detailed documentation}
\label{appendix:musicorpus}

MusiCorpus defines a set of guidelines on how to structure an Optical Music Recognition (OMR) dataset. 
The goal is to 
decouple OMR data consumers from the specifics of how the dataset they use is structured. This will allow for easy replacement of datasets in experiments as well as their combination into larger corpora. Ideas here described build on the work by Shatri \textit{et al.} \cite{shatriDoReMiFirstGlance2021} in an attempt to make new datasets easier to compose together.

\subsection{Directory structure}

The directory structure for MusiCorpus data is the following:

\begin{verbatim}
root/                          (root folder for everything)
|
+-- CVC.Dolores/               (dataset - a list of pages)
|   +-- musicorpus.json        (metadata about the dataset)
|   +-- README.md              (dataset description)
|   +-- LICENSE.txt            (license for the data)
|   +-- splits.json            (train/val/test splits)
|   |
|   +-- CEDOC_CMM_1.5.1_0081.001/     (page name)
|   |   +-- image.jpg
|   |   +-- metadata.json
|   |   +-- subdivisions.image.json
|   |   +-- subdivisions.coco-object-detection.json
|   |   +-- layout.json
|   |   +-- coco-object-detection.json
|   |   +-- transcription.musicxml
|   |   +-- transcription.krn
|   |   +-- transcription.mei
|   |   +-- transcription.ly
|   |   +-- transcription.midi
|   |   +-- transcription.mung
|   |   |
|   |   +-- Systems/
|   |   |   +-- 1/
|   |   |   |   +-- image.jpg
|   |   |   |   +-- transcription.musicxml
|   |   |   |   `-- ...
|   |   |   +-- 2/
|   |   |   `-- 3/
|   |   |
|   |   +-- Staves/
|   |   |   +-- 1/
|   |   |   +-- 2/
|   |   |   `-- ...
|   |   |
|   |   `-- Grandstaves/
|   |
|   |   (more pages)
|   +-- UAB_LICEU_222570.112/
|   +-- CEDOC_CMM_1.5.1_0081.002/
|   `-- XAC_ACUR_TagFAu127_007/
|
+-- UFAL.OmniOMR/                   (another dataset)
|   +-- ...                         (its top-level files)
|   |
|   |   (pages)
|   +-- 1d507bc2-87e7-4b...c2-80a6-cccbcaeb7893/
|   +-- 6aa42890-26ea-11...49-8dd8-ae38fe6f4744/
|   +-- 11ccf60d-cc2e-48...25-a986-f95d083e6142/
|   `-- ...
|
`-- YourInstitution.YourDataset/
    `-- ...
\end{verbatim}

\subsection{Dataset Folder}

MusiCorpus aids dataset harmonization between researchers by providing guidelines on a shared folder structure and its semantics. However each dataset creator is still responsible for publishing their own dataset, keeping it up to date and following the guidelines. That's why each MusiCorpus-compatible dataset is released as a separate, independent folder. These folders are then placed into the root directory for ease of consumption by the user.

The dataset folder's name must provide the name of the organization that packaged the dataset and the name of the dataset following this pattern:

\begin{verbatim}
    [Organisation Name].[Dataset Name]
\end{verbatim}

Both the organisation name and dataset name must only include alphanumeric characters with CapCase spelling. We recommend using abbreviated forms for both. For example, the Computer Vision Center-built DoLoReS dataset can be found under the \verb`CVC.Dolores` folder.

A dataset is conceptualized as a set of pages, where each page image comes with machine-readable annotations. This page-primacy comes from the way music notation is physically realised. Libraries hold books with pages of music notation and notation software (e.g. MuseScore) produces PDFs with pages of music notation.

Files in the dataset folder contain dataset-level metadata and folders represent individual pages of music notation (see \autoref{sec:PageFolders}).

Despite the folder structure being page-organized you are not required to provide page-level data in your dataset to comply with MusiCorpus. Each page may contain so-called \textit{subdivisions}, such as systems, staves and grandstaves. Their structure is equivalent to page-folders, but they focus on smaller subdivisions of music notation. Subdivisions can be provided in addition to page-level data (as crop-outs from pages) or completely stand alone with the page-level data missing.

\paragraph{Example 1} The OmniOMR dataset primarily contains pages of music notation annotated in MusicXML and MuNG and then contains subdivisions to systems on the page, where each system contains a cropped version of the page image, cropped version of the MuNG annotaion and a sliced out version of the MusicXML annotation. Here, the system-level subdivision is a different view of the same page-level data.

\paragraph{Example 2} The DoLoReS dataset is annotated as system-level images and system-level MusicXML transcriptions. The page-level annotations are then created ex post by concatenating the system-level MusicXML files.

\paragraph{Example 3} The OLiMPiC scanned dataset \footnote{\url{https://github.com/ufal/olimpic-icdar24/releases/tag/datasets}} \cite{mayerPracticalEndToEnd2023} only contains grandstaff-level MusicXML annotations and grandstaff image crops. These come from pages, but the original page scans are not present in the dataset. It can be reshaped into the MusiCorpus structure in this way:

\begin{verbatim}
UFAL.OlimpicScanned/
+-- musicorpus.json
+-- ...
|
+-- 5026306-p5/
|   `-- Grandstaves/
|       +-- 1/
|       |   +-- image.jpg
|       |   `-- transcription.musicxml
|       `-- 2/
|           +-- image.jpg
|           `-- transcription.musicxml
+-- 5026306-p6/
+-- ...
+-- 6377942-p1/
`-- 6377942-p2/
\end{verbatim}

Notice that OLiMPiC is organized into "scores" (e.g. \verb`5026306`) and that score has samples \verb`p5-s1.png` (page 5, grandstaff 1). In MusiCorpus these are re-grouped into page names \verb`5026306-p5` (score 5026306, page 5) and that page has no page-level information, only grandstaff-level image and MusicXML transcription.

\subsection{musicorpus.json}

A \verb`musicorpus.json` must sit in the root of every MusiCorpus-compatible dataset. It provides metadata about the whole dataset. This is an example \verb`musicorpus.json` file for the \verb`CVC.Dolores` dataset:

\begin{lstlisting}[language=Java]
{
    "musicorpus_version": "1.0",
    "full_institution_name": "Computer Vision Center, Universitat Autonoma de Barcelona",
    "short_institution_name": "CVC",
    "institution_url": "https://www.cvc.uab.es/",
    "full_dataset_name": "DoLoReS",
    "short_dataset_name": "Dolores",
    "dataset_url": "https://pages.cvc.uab.es/musicscores/loladocs/index.html",
    "dataset_version": "1.0",
    "created_at": "2026-03-05T10:16:37Z",
    "author_emails": [
        "ptorras@cvc.uab.cat",
        "afornes@cvc.uab.es"
    ]
}
\end{lstlisting}

\begin{itemize}[noitemsep,leftmargin=*]
    \item \verb`musicorpus_version`: Version of the MusiCorpus dataset format used, as well as the version of the \verb`musicorpus.json` file format.
    \item \verb`full_institution_name`: Human-readable name of the institution behind the dataset. If only an individual person, then the full name of the individual.
    \item \verb`short_institution_name`: The first-half of the dataset folder name (i.e. \verb`*CVC*.Dolores`), must match exactly and be a path-safe string.
    \item \verb`institution_url`: URL link to the website of the institution. If an individual person, link to their website or may be left empty.
    \item \verb`full_dataset_name`: Human-readable name of the dataset.
    \item \verb`short_dataset_name`: The second-half of the dataset folder name (i.e. \verb`CVC.*Dolores*`), must match exactly and be a path-safe string.
    \item \verb`dataset_url`: URL link to the website about the dataset or the project from which the dataset arose. May be left empty.
    \item \verb`dataset_version`: Version of the dataset in the \verb`{major}.{minor}` format. Increment \verb`minor` whenever you fix bugs in the dataset or add additional transcriptions or subdivisions (systems, staves, grandstaves) of existing data. Increment \verb`major` when you change dataset splits, add/remove data samples or change versions or semantics of existing transcriptions. During development, you can use \verb`0.x` versions and increment \verb`minor` with any changes as you see fit.
    \item \verb`created_at`: ISO 8601 timestamp of the moment the dataset was put together (exported by a script or considered "done" by a human).
    \item \verb`author_emails`:  List of email addresses of authors, sorted by who should be emailed first/second/etc.
\end{itemize}

\subsection{README.md}

The \verb`README.md` file is the first entrypoint for the user of a MusiCorpus dataset. It should contain all the non-structured information about your dataset a user may wish to know. Below is a list of questions that the \verb`README.md` file should answer.

\begin{itemize}[noitemsep,leftmargin=*]
    \item What task is the dataset intended for?
    \item How much data is included in the dataset? Number of pages, staves, etc.
    \item What is the source data?
    \item What data was manually annotated?
    \item What data was derived computationally?
    \item Is there a structure behind page names?
    \item Are pages primary and systems/staves/grandstaves derived or is it the over way around?
    \item What version of MuseScore was used to annotate the MusicXML, what semantics was used etc... (analogously for other annotation formats)
    \item Does the dataset use some annotation format in some non-obvious way? (e.g. MuNG only for bboxes, but not for masks)
    \item Which files are present in all pages/systems/staves and which only in some?
    \item Are the source images accessible via the internet? At what URLs?
    \item Does the dataset deviate from the MusiCorpus guidelines? Where and how?
    \item What is the license/licenses for the data?
    \item What files are under what license?
\end{itemize}

\subsection{LICENSE.txt}

This file contains the legal code of the license, under which the dataset is published. If there are multiple licenses used (for example, the images have different license than the transcriptions or a set of pages has a different license etc.), provide multiple license files with this naming:

\begin{verbatim}
LICENSE.txt                 (the default license)
LICENSE.images.txt          (license for images)
LICENSE.transcriptions.txt  (license for transcriptions)
LICENSE.my-split.txt        (license for pages in "my-split")
\end{verbatim}

When multiple licenses are provided, it must be specified in the \verb`README.md` file which licenses apply to which files in the dataset.

\subsection{splits.json}

Each dataset consists of a list of pages, and these pages are cut into splits designed for model training, validation and testing. The \verb`splits.json` file defines which pages belong to which split.

Make sure your \verb`README.md` contains information on how these splits were generated (if there are any independence guarantees, such as handwriting author, or if it is just a random shuffle).

This is the internal structure of a \verb`splits.json` file:

\begin{lstlisting}[language=Java]
{
    "train": [
        "CEDOC_CMM_1.5.1_0081.001",
        "CEDOC_CMM_1.5.1_0081.002",
        ...
    ],
    "validation": [
        "XAC_ACUR_TagFAu127_007",
        ...
    ],
    "test": [
        "UAB_LICEU_222570.112",
        ...
    ],
}
\end{lstlisting}

The three splits in \verb`splits.json` must not share any pages - they must be disjoint sets. They are not required to cover the complete set of pages (say if you want to make the splits compatible with alternative splits), but it is recommended. Also, the \verb`validation` set is optional, but recommended.

If you want to provide multiple different splits (e.g. to ensure different independencies or use-cases), you can do so by naming these alternative splits in this way (see below). Make sure to also document their purpose in the \verb`README.md` file.

\begin{verbatim}
splits.writer-independent.json
splits.domain-adaptation.json
splits.{your-splits-name}.json
\end{verbatim}

In these alternative splits, you can add additional sets of pages, say \verb`holdout` or \verb`finetuning` if it makes sense. But the primary \verb`splits.json` file should be left in its default, simple structure of train-validation-test. You can create an alternative split, such as \verb`splits.with-holdout.json` for your additional page sets.

\subsection{Page Folders}
\label{sec:PageFolders}

As mentioned, each folder within a dataset corresponds to all of the data associated with a page. The naming scheme for page folders is up to the dataset creator to decide. The page name structure should be described in the \verb`README.md`, if there is any.

Each page folder contains files that hold data for that page, such as:

\begin{itemize}[noitemsep,leftmargin=*]
    \item Image (scan) of the page (JPEG)
    \item Page-level metadata (page source information, notation complexity)
    \item Object detection annotations (COCO)
    \item Transcriptions to MusicXML, kern, MEI, abc, etc.
\end{itemize}

Until recently, there were no page-level end-to-end deep learning models. Only one-staff and then one-grandstaff models were available. However, given that research on single-staff modeling remains incomplete and that training page-level models introduces significant complexity, MusiCorpus provides subdivisions - a narrowed view at a page at the staff level. These subdivisions are organized within a page-specific directory, which may optionally include up to three division-level subfolders structured as follows:

\begin{itemize}[noitemsep,leftmargin=*]
    \item \verb`Staves` (for solo-staff models)
    \begin{itemize}
        \item \verb`1/`, \verb`2/`, \verb`3/`, ...
    \end{itemize}
    \item \verb`Grandstaves` (for piano grandstaff models)
    \begin{itemize}
        \item \verb`1-2/`, \verb`3-4/`, ...
    \end{itemize}
    \item \verb`Systems` (for system-level models)
    \begin{itemize}
        \item \verb`2-7/`, \verb`9-14/`, \verb`15-20/`, ...
    \end{itemize}
\end{itemize}

Each may contain a list of zoomed-in views of the page, to all of its systems, staves or grandstaves. Each subdivision folder then may contain the same image and transcription files as the page folder, just zoomed in and cropped respectively. Names of systems, staves and grandstaves can be any path-safe strings, however, MusiCorpus recommends using the following policy when possible.

\begin{figure}[h]
    \centering
    \includegraphics[width=0.5\linewidth]{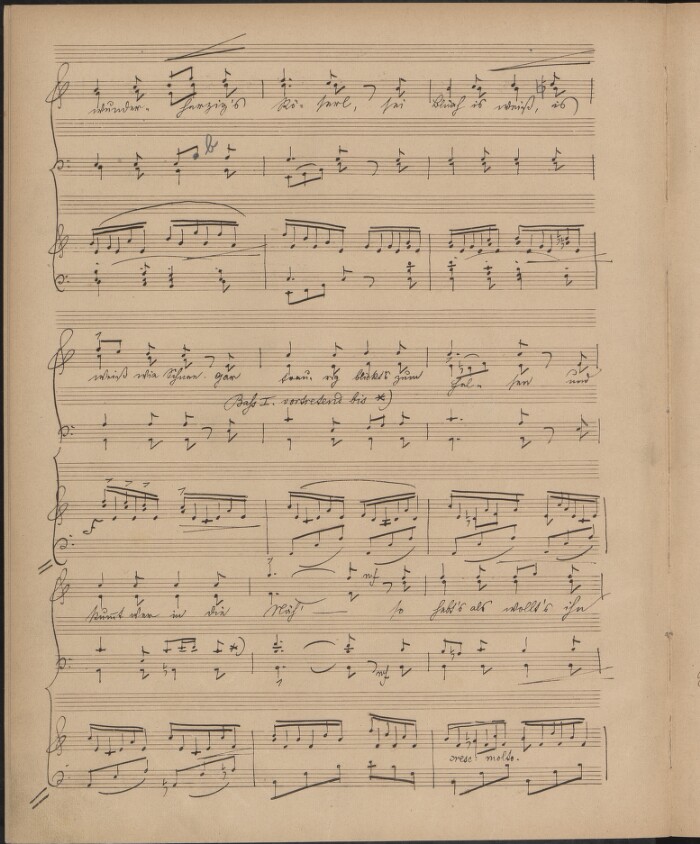}
    \caption{An example manuscript page with interesting staff, grandstaff, and system composition. It can be viewed online at \url{http://digitalniknihovna.cz/mzk/uuid/uuid:d1769738-290b-4810-90b7-19fd8708d0c7}.}
    \label{fig:mzkPageD17}
\end{figure}

Take the page in \autoref{fig:mzkPageD17} as an example. There are 20 staves on the page, we number them \verb`1` to \verb`20` (1-based index). For the \verb`Staves` subdivision, we could take all 20 staves and use them as is, but the staves \verb`1,3,5,8,10,12,18` are empty and contain no music. If we have the information available, we can discard these and only create subdivision folders for the remaining staves:

\begin{verbatim}
8136b106-6283-42c6-99...b7-19fd8708d0c7/
`-- Staves/
    +-- 2/
    |   +-- image.jpg
    |   `-- transcription.musicxml
    `-- 4/, 6/, 7/, 9/, 11/, 13/, 14/, 15/, 16/, 17/
\end{verbatim}

\begin{figure}[h]
    \centering
    \includegraphics[width=0.9\linewidth]{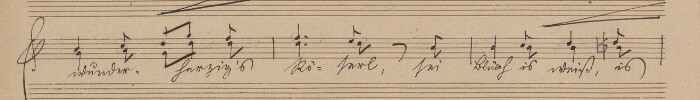}
    \caption{The staff 2 image crop from the page in \autoref{fig:mzkPageD17}.}
    \label{fig:mzkPageD17-staff2}
\end{figure}

The \verb`Staves/2/image.jpg` file is shown in \autoref{fig:mzkPageD17-staff2}. The \verb`Staves` subdivision is meant for training solo-staff models. The notation complexity may be arbitrarily complex (monophonic or polyphonic), but it should be contained on a single staff.

The \verb`Grandstaves` subdivision is an analogous subdivision meant for single-grandstaff (piano) end-to-end models. It should always contain 2 staves, where the notation on them resembles piano music (either the staves are grouped by a brace, or they belong together based on the G-clef, F-clef combo). Grandstaff folders should be composed from the staff numbers defined above:

\begin{verbatim}
8136b106-6283-42c6-99...b7-19fd8708d0c7/
`-- Grandstaves/
    +-- 6-7/
    |   +-- image.jpg
    |   `-- transcription.musicxml
    +-- 13-14/
    `-- 19-20/
\end{verbatim}

If you have a dataset, where you can't resolve the staff numbers, you can just number the grandstaves from \verb`1` and increasing (\verb`1/`, \verb`2/`, \verb`3/`).

\begin{figure}[h]
    \centering
    \includegraphics[width=0.9\linewidth]{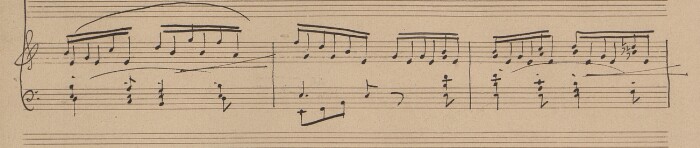}
    \caption{The grandstaff 6-7 image crop from the page in \autoref{fig:mzkPageD17}.}
    \label{fig:mzkPageD17-grandstaff67}
\end{figure}

\begin{figure}[h]
    \centering
    \includegraphics[width=0.9\linewidth]{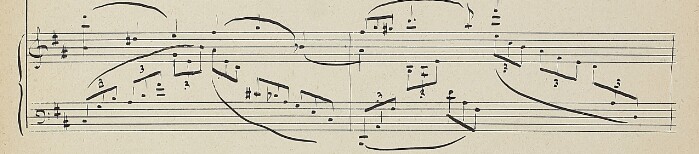}
    \caption{An example of a grandstaff that cannot be represented as two separate staves due to the pianoform music notation present on it.}
    \label{fig:mzkPage308-grandstaff}
\end{figure}

The \verb`Grandstaves/6-7/image.jpg` file is shown in \autoref{fig:mzkPageD17-grandstaff67} Note that a single staff may be present in both the \verb`Staves` subdivision and the \verb`Grandstaves` subdivision as a part of a grandstaff. This is, however, only possible for grandstaves that contain staff-separable music. In complex pianoform music, where beams and voices cross staves of the grandstaff, that grandstaff may only be present in \verb`Grandstaves` but cannot be present in \verb`Staves`, because the music notation on one staff does not remain only on that staff. An example of a grandstaff, that cannot be separated out into individual staves can be seen in \autoref{fig:mzkPage308-grandstaff}.

Finally the \verb`Systems` subdivision is again analogous to \verb`Grandstaves`, except it captures individual systems (staves of all instruments that play together). The system folders should again be composed of staff numbers from above:

\begin{verbatim}
8136b106-6283-42c6-99...b7-19fd8708d0c7/
`-- Systems/
    +-- 2-7/
    |   +-- image.jpg
    |   `-- transcription.musicxml
    +-- 9-14/
    `-- 15-20/
\end{verbatim}

If you have a dataset, where you can't resolve the staff numbers, you can just number the systems from \verb`1` and increasing (\verb`1/`, \verb`2/`, \verb`3/`).

\subsection{image.jpg}

In each page, staff, grandstaff or system, there may be the \verb`image.jpg` file which contains the image being recognized and transcribed.

If an image is to be made available, it must be provided in the \verb`.jpg` suffix. The \verb`.jpeg` suffix is not recommended. In addition, you may provide \verb`.png` or \verb`.tif` images if your image is born-digital and you want to preserve its quality.

\subsection{subdivisions.image.json}

If you have the page-level \verb`image.jpg` files and also staff/grandstaff/system level \verb`image.jpg` files, these smaller images are just simple rectangular crops of the larger page-level image. The \verb`subdivisions.image.json` file can be placed in the page folder to specify the exact coordinates, where the subdivision images were cropped from the page-level image. This is the structure of the file:

\begin{lstlisting}[language=Java]
{
    "Staves": {
        "1": {
            //      left  top  width height   (COCO-style)
            "bbox": [151, 112, 2601, 240]
        },
        ...
    },
    "Grandstaves": {
        "6-7": {
            "bbox": [162, 908, 2572, 456]
        },
        ...
    },
    "Systems": {}
}
\end{lstlisting}

For each subdivision image there is a JSON object at a corresponding "path". For \verb`Grandstaves/6-7/image.jpg`, the JSON object is at \verb`"Grandstaves"."6-7"`. The JSON object has just a single field, and that is \verb`bbox`:

\begin{verbatim}
{
    "bbox": [162, 908, 2572, 456]
}
\end{verbatim}

The \verb`bbox` field is a COCO-style rectangle \verb`[left, top, width, height]` where \verb`left` and \verb`top` is 0-based pixel index of the top-left pixel of the rectangle and \verb`width` and \verb`height` are the dimensions of the rectangle in pixels. This should exactly match the size of the subdivision \verb`image.jpg` file.

All pixel coordinates are the parent page-level \verb`image.jpg` pixel coordinates (i.e. they can be used directly as-is to crop out the small subdivision image).

These coordinates locate the subdivided image in the page-level image, they do not locate the bounding box of a staff, grandstaff or system. For precise locations of layout-features see the \verb`layout.json` file.

\subsection{metadata.json}

Each page in the dataset should contain a metadata file which describes the source document itself (the image / piece of paper), the musical work in that page (composer, date) and music notation information, such as its complexity (monophonic vs. polyphonic, etc.), type of notation (CMWN, mensural, etc.), production (handwritten/printed/born-digital), clarity/readability, etc.

The metadata is stored as a list of fields of a JSON file. If a value is unknown for the given page, \verb`null` should be used. If a field does not make sense for the given page, \verb`false` should be used (e.g. author name for synthetically generated music). If a field is missing all together, it is assumed to have the value of \verb`null`. This is an example record:

\begin{lstlisting}[language=Java]
{
    "file_name": "CVC.Dolores/CEDOC_CMM_1.5.1_0081.001/image.jpg",

    // source document metadata
    "institution_name": "Moravian State Library",
    "institution_rism_siglum": "CZ-Bu",
    "institution_local_siglum": "MZK",
    "shelfmark": null,
    "rism_id_number": "sources/1001086460",
    "date": 1804,
    "page_number": 14,
    "page_size": [235, 285],
    "dpi": 275,
    "scribal_data": null,
    "link": "https://www.e-rara.ch/mab/content/zoom/28920816",

    // musical work metadata
    "title_description": "Sonata I [G-Dur op. 40 Nr. 1]",
    "author": "Muzio Clementi",
    "author_date": "1752-1832",
    "genre_form": "piano sonata",
    
    // music notation and OMR metadata
    "notation": "CWMN",
    "notation_detailed": null,
    "production": "printed",
    "production_detailed": "19th century engraving, highly standardised",
    "notation_complexity": "pianoform",
    "clarity": "perfect",
    "systems": "grand_staffs"
}
\end{lstlisting}

This is the meaning of individual fields:

\begin{itemize}[noitemsep,leftmargin=*]
    \item \verb`file_name`: Path to the image (relative to the \verb`root/` folder) this metadata file describes.
\end{itemize}

Then come fields related to the source:

\begin{itemize}[noitemsep,leftmargin=*]
    \item \verb`institution_name`: Human-readable name of the holding institution (library). Should be \verb`false` for images that are born-digital and are not held by any institution.
    \item \verb`institution_rism_siglum`: RISM\footnote{\url{https://rism.info/community/sigla.html}} assigns abbreviations to institutions holding musical works. This field should contain the siglum of the institution name above. Should be \verb`null` if the institution does not have a siglum yet. Should be \verb`false` if the source document is not held in an institution.
    \item \verb`institution_local_siglum`: The abbreviation by which the institution refers to itself. Should be \verb`false` if the source document is not held in an institution.
    \item \verb`shelfmark`: Identifies the source (manuscript, print, ...) within the institution's collection. Same as RISM shelfmark. Should be \verb`null` if unknown. Should be \verb`false` if the source document is not held in an institution.
    \item \verb`rism_id_number`: If the source is catalogued in RISM, then this should be given. This field acts as a fallback if \verb`shelfmark` does not exist. Should be \verb`false` if not catalogued or \verb`null` if unknown.
    \item \verb`date`: The year when the source was created. This can be very different from when the musical piece within was composed (e.g. 20th century editions of Renaissance music). A specific year is usually very hard to pinpoint so instead this field is a string which allows for best-effort description of the year  range if a singular year cannot be pinned down. Should be \verb`null` if unknown.
    \item \verb`page_number`: Some identification of the image within the source. Can be just a number, or foliation (e.g. \verb`f55v`). Intended to be human-readable. Should be \verb`null` if unknown.
    \item \verb`page_size`: Size of the physical page/book in millimeters, \verb`[w, h]`. It may be used to estimate DPI if DPI is not explicitly provided (with estimation error given by the margin around the page in the page scan image). Should be \verb`null` if unknown.
    \item \verb`dpi`: Actual DPI at which the image was scanned. Ideally as precise as possible - estimated via color calibration tables or rulers. Should be \verb`null` if unknown.
    \item \verb`scribal_data`: Identifier of a handwriting/typesetting style (string or integer). May be used by a dataset intended for writer identification. Should be \verb`null` if unknown or if not provided. Identifiers should be unique within the dataset.
    \item \verb`link`: URL to the page image, if available. A URL from the holding institution is preferred. If not available, then another digitised version in RISM. Should be \verb`null` if unknown or if such a link does not exist.
    \item \verb`title_description`: Name of the composition (the musical piece captured within the document), e.g. "Sonata XYZ". Stick to RISM as much as possible. Should be \verb`null` if unknown. Should be \verb`false` if not aplicable (e.g. synthetic music).
    \item \verb`author`: Name of the composer, should be \verb`null` if unknown. Should be \verb`false` if not aplicable (e.g. synthetic music). Stick to RISM naming.
    \item \verb`author_date`: Lifespan of the author, range of integers, e.g. \verb`1752-1832`. Use \verb`?` if one of the dates is uncertain, e.g. \verb`?-1832`. Should be \verb`null` if fully unknown. Should be \verb`false` if not aplicable (e.g. synthetic music). Stick to RISM dating system.
    \item \verb`genre_form`: Human-readable, uncontrolled characterisation of the work, e.g. \verb`piano sonata`, \verb`sting quartet`, \verb`motet`. Should be \verb`null` if not provided.
\end{itemize}

Finally, and most importantly for OMR, there are fields that characterise the notation on the page. Controlled vocabularies build on the taxonomy presented in the "Understanding OMR" paper \cite{calvo-zaragozaUnderstandingOpticalMusic2021}.

\begin{itemize}[noitemsep,leftmargin=*]
    \item \verb`notation`: Type of music notation. Must be one of: \verb`CWMN`, \verb`mensural`, \verb`square`, \verb`adiastematic`, \verb`instrument-specific`, \verb`other`. Should be \verb`null` if unknown or not specified.
    \item \verb`notation_detailed`: An optional fine-grained description, such as \verb`Black` \verb`menusral`, \verb`Suzipu`, \verb`Franconian`, \verb`German 16th c. lute tablature`, \texttt{Jeonggangbo}, \verb`lead sheet`, \verb`mountain hymnals`, etc. Should be \verb`null` if not specified.
    \item \verb`notation_complexity`: One of the four levels defined in the "Understanding OMR" paper. Must be one of: \verb`monophonic`, \verb`homophonic`, \verb`polyphonic`, \verb`pianoform`. Mark the page according to the most complex level that it contains -- for instance, the score of a piano concerto with just one grand staff with pianoform notation and everything else monophonic will be marked \verb`pianoform`. The point is to signal "in order to recognise this page completely, one must account for this level of notation complexity". Should be \verb`null` if not specified.
    \item \verb`production`: How was the music document created. Must be one of: \verb`printed`, \verb`handwritten`, \verb`born-digital`. Should be \verb`null` if not specified. Printed includes old printed scores as well as modern ink-jet printed and then scanned scores. Born-digital means being rendered by Verovio, Lilypond, MuseScore and other similar software directly to a JPEG/PNG file.
    \item \verb`production_detailed`: Optional more fine-grained description, e.g.: \verb`Muse` \verb`Score`, \verb`LilyPond with font`, \verb`movable type`, \verb`sharpie marker`, \verb`quill`, etc. Should be \verb`null` if not specified.
    \item \verb`clarity`: The combined visual quality/readability of the page. There are four categories: \verb`perfect`, \verb`sufficient`, \verb`problematic`, and \verb`unreadable`. More discussion on this category, including criteria, below. Should be \verb`null` if not specified.
    \item \verb`systems`: Basic information about the relationship between staffs, systems, and reading order. Again, it serves as a signpost: "What do I have to care about when I use this dataset?". The field value is the answer to the question: "What does a single system of music consists of in this document?". Must be one of: \verb`single-staff`, \verb`grand-staff`, \verb`multi-instrument`, \verb`variable`. Should be \verb`null` if not specified. If all systems on a page are of one type, then that type is the value of this field. If there is at least one system of different type (e.g. when there are multiple songs within one page with different setup), then \verb`variable` should be used.
\end{itemize}

\paragraph{The \textit{clarity} field} This categorisation provides a rough idea of the kinds of issues connected to image quality, document damage, readability, and other difficulties connected to visual aspects of the notation one might encounter on the page. It is not intended to be an exhaustive scale, just a rough indication. The point of this taxonomy is also to give insight into how robust systems are to this kind of difficulties when evaluating them (accuracy on category 1 pages, category 2, etc.).

\begin{enumerate}[noitemsep,leftmargin=*]
    \item \verb`perfect`: Born-digital or scan/photo with good even lighting. No bleedthrough, stains, scribbles nor deformations.
    \item \verb`sufficient`: Does not fulfill conditions of category 1, yet remains readable at sight by an experienced musician without significant difficulty. (For manuscripts, consider performers experienced in historically informed performance practice.)
    \item \verb`problematic`: Bad handwriting, damage that interferes with ability to sight-read, requires time. Impractical --- a musician might refuse to play from this music. But with sufficient effort the music could be transcribed unambiguously from it, just some isolated elements may be ambiguous or unreadable.
    \item \verb`unreadable`: The music is difficult to read to the point of ambiguity ("Which line is this notehead on? Is this a staccato dot or an artifact? What notes does this beam belong to?"), page is heavily damaged, imaging quality is terrible (blur or bad/uneven lighting to the point of damaging readability), etc., and these issues are pervasive (affect more than 5-10\% of the music on the page surface).
\end{enumerate}

This is just a way to explain what the categories should approximately mean. "What issues you have to care about when you're doing OMR system design over this dataset." The point of this is also to have training data for identifying problematic pages in collections that will need manual attention.

\subsection{coco-object-detection.json}

Each page, system, staff, or grandstaff may contain a \verb`coco-object-detection` \verb`.json` file that contains MS COCO object detection annotations\footnote{\url{https://cocodataset.org/\#format-data}} for the corresponding \verb`image.jpg` file (e.g. for the COCO file in a staff folder, the annotations correspond to the image in the same staff folder). This is a shortened example for the page \verb`CEDOC_CMM_1.5.1_0081.001` in the \verb`CVC.Dolores` dataset:

\begin{lstlisting}[language=Java]
{
    "info": {
        "year": 2025,
        "version": "1.0",
        "description": "CVC.Dolores",
        "contributor": "Computer Vision Center, Universitat Autonoma de Barcelona",
        "url": "https://pages.cvc.uab.es/musicscores/loladocs/index.html",
        "date_created": "2025/11/28"
    },
    "licenses": [
        {
            "id": 0,
            "name": "CVC.Dolores/LICENSE.txt",
            "url": "musicorpus://CVC.Dolores/LICENSE.txt",
        }
    ],
    "images": [
        {
            "id": 0,
            "width": 3364,
            "height": 4010,
            "file_name": "CEDOC_CMM_1.5.1_0081.001/image.jpg",
            "license": 0,
            "date_captured": "2025-11-28 13:23:25"
        }
    ],
    "annotations": [
        {
            "id": 0,
            "image_id": 0,
            "category_id": 0,
            "segmentation": [[356, 428, 357, 428, ...]], // polygon
            "area": 546,
            "bbox": [641, 471, 21, 26],
            "iscrowd": 0
        },
        {
            "id": 1,
            "image_id": 0,
            "category_id": 4,
            "segmentation": { // RLE mask
                "size": [24, 20],
                "counts": [8, 2, 9, 1, 4, ...]
            },
            "area": 480,
            "bbox": [679, 445, 20, 24],
            "iscrowd": 0
        },
        ...
    ],
    "categories": [
        { "id": 0, "name": "noteheadBlack" },
        { "id": 1, "name": "accidentalFlat" },
        { "id": 2, "name": "barlineSingle"},
        { "id": 3, "name": "beam"},
        { "id": 4, "name": "fermataAbove"}
        ...
    ]
}
\end{lstlisting}

\begin{itemize}[noitemsep,leftmargin=*]
    \item \verb`info`: An object that contains dataset-level metadata. This value should be identical for all \verb`coco-object-detection.json` files in the dataset.
    \begin{itemize}
        \item \verb`year`: Must equal the year value of the \verb`date_created` field below.
        \item \verb`version`: Version of the dataset. Must equal the \verb`musicorpus.json/` \verb`dataset_version` field.
        \item \verb`description`: Name of the MusiCorpus dataset. Must equal the dataset folder name as well as the \verb`musicorpus.json` fields \verb`short_` \verb`institution_name` and \verb`short_dataset_name`.
        \item \verb`contributor`: Name of the institution or the individual behind the dataset. Must equal to \verb`musicorpus.json/full_institution_name` field.
        \item \verb`url`: URL link to a website about the dataset or project. Must equal the \verb`musicorpus.json/dataset_url` field.
        \item \verb`date_created`: Date when the dataset was created. Must have format \verb`YYYY/MM/DD` and be equal to the value of the \verb`musicorpus.json/` \verb`created_at` field.
    \end{itemize}
    \item \verb`licenses`: List of licenses used in this COCO file.
    \begin{itemize}
        \item \verb`id`: Integer ID of the license within this COCO file.
        \item \verb`name`: Human-readable name of the license, such as \verb`CC BY-SA 4.0`, however for simplicity, you can set this field to the MusiCorpus path to the license file, e.g. \verb`CVC.Dolores/LICENSE.txt` or if a specific license is used then \texttt{CVC.Dolores/LICENSE.transcriptions.txt}, etc.
        \item \verb`url`: URL link to the license body. Because a MusiCorpus dataset contains the license file within itself, the link should look like this: \texttt{musicorpus://CVC.Dolores/LICENSE.txt}. The custom scheme is there to differentiate it from HTTP links. However HTTP links may also be used, if your license is hosted on a website and you prefer this option.
    \end{itemize}
    \item \verb`images`: Contains the single image for which this \verb`coco-object-detection` \verb`.json` file holds annotations.
    \begin{itemize}
        \item \verb`id`: Integer ID of the image within this COCO file.
        \item \verb`width`: Width of the image in pixels.
        \item \verb`height`: Height of the image in pixels.
        \item \verb`file_name`: Posix path (forward slashes) from the root of the dataset (\verb`CVC.Dolores`) to the image file.
        \item \verb`license`: The license ID that applies to the image file.
        \item \verb`date_captured`: Timestamp when the image file was first created. This can be the time it was scanned, or created (for born-digital images). If unavailable, it can be set to the creation time of the dataset (the \verb`musicorpus.json/created_at` field). It must have the \verb`YYYY-MM-DD hh:mm:ss` format and be in UTC time.
    \end{itemize}
    \item \verb`anotations`: List of annotated object instances in the image.
    \begin{itemize}
        \item \verb`id`: Integer ID of the object instance within this COCO file. If the same object is present in two overlapping images (say a page and its staff subdivision), the mapping between their COCO IDs is captured in the \verb`subdivisions.coco-object-detection.json` file in the page folder.
        \item \verb`image_id`: What image does this annotation belong to (see `images` above).
        \item \verb`category_id`: What is the "label" (e.g. notehead black) for this object (see \verb`categories` below).
        \item \verb`segmentation`: Polygon or RLE-encoded pixel mask for the object.
        \item \verb`area`: Number of pixels in the annotated area (not just bbox area, but actual mask area).
        \item \verb`bbox`: Bounding box for the mask in the form \verb`[x, y, w, h]` where all numbers are integers and x and y are zero-based coordinates.
        \item \verb`iscrowd`: Determines, whether the annotation contains a single instance (\verb`0`) or s crowd of instances (\verb`1`). For music notation glyphs it makes sense to only annotate individual instances, so this field should always be set to \verb`0`.
    \end{itemize}
    \item \verb`categories`: Mapping between category ("label") IDs and names. The mapping contains only those categories that are actually used inside the \verb`annotations` list above.
    \begin{itemize}
        \item \verb`id`: Integer ID of the category within this COCO file. One category name must always have exactly one ID (no duplicate categories).
        \item \verb`name`: String name of the category, e.g. \verb`noteheadBlack`.
        \item \verb`supercategory`: Not used in MusiCorpus, despite COCO allowing it.
    \end{itemize}
\end{itemize}

The \verb`segmentation` can be represented as polygons, or as RLE-encoded pixel masks. \textbf{Polygon representation} consists of of a list of polygons, where a polygon is a list of 2D points. These are valid polygon segmentations:

\begin{lstlisting}[language=Java]
{
    // single polygon - a quadrangle with points A B C D
    "segmentation:" [
        [Ax, Ay, Bx, By, Cx, Cy, Dx, Dy]
    ],

    // a quadrangle and a triangle, both representing one object
    "segmentation:" [
        [Ax, Ay, Bx, By, Cx, Cy, Dx, Dy],
        [Ex, Ey, Fx, Fy, Gx, Gy]
    ],
}
\end{lstlisting}

\textbf{Uncompressed RLE representation} consists of a JSON dictionary which looks like this:

\begin{lstlisting}[language=Java]
{
    // 12x5 mask with 2px wide vertical stripes
    "segmentation": {
        //       h   w
        "size": [5, 12],
        //         0s  1s  0s  1s  0s  1s
        "counts": [10, 10, 10, 10, 10, 10]
    }
}
\end{lstlisting}

The \verb`size` is \verb`[height, width]` and must equal the \verb`bbox` size of the annotation. The \verb`counts` field is a list of runs of zeros and ones, where zero means no mask and one means mask is present. The first run is a run of zeros, therefore all odd runs are zeros and even runs are ones.

To remove friction when combining OMR object-detection datasets, MusiCorpus requires \verb`coco-object-detection.json` files to use the \href{https://github.com/OmniOMR/mung/blob/7bddc87d61e19b62ac46834a66c88239dbfebdc5/docs/annotation-instructions/annotation-instructions.md}{MuNG 2.0} naming and semantics for music notation objects. The MuNG 2.0 ontology builds on top of \href{https://www.smufl.org/}{SMuFL} and naturally continues the tendency for class name unification in the OMR field.

If your annotation categories have different names, please introduce a mapping when exporting your dataset to MusiCorpus and align your names to SMuFL and MuNG 2.0 as closely as possible. For edgecases that are impossible to resolve without data re-annotation, try to use a sensible mapping that minimizes confusion for object detection models trained on the data.

Make sure to explain in the dataset \verb`README.md` file how closely your data aligns to the MuNG 2.0 and SMuFL semantics and list important deviations. If you use completely different semantics, describe it in the \verb`README.md` also.

If you want to annotate \textit{regions} in the image (e.g. staves, background, lyrics, notes), basically doing semantic segmentation instead of instance segmentation, use \verb`coco-stuff-segmentation.json` files instead. MusiCorpus currently lacks any opinions on that file structure or semantics and names of categories. Define this yourself as a reasonable extension of the MusiCorpus format so that it may be included in a future MusiCorpus version.

\subsection{subdivisions.coco-object-detection.json}

Just like the \verb`subdivisions.image.json` file maps subdivision images to the page-level image via crop bounding box coordinates, this file maps COCO annotation IDs. When the same physical symbol on the physical page is captured in both the page-level COCO annotation and a staff-level subdivision annotation, the \verb`subdivisions.coco-object-detection.json` file maps the file-local COCO IDs of the annotation instances (\verb`"annotations"."id"`). This is what the mapping file looks like:

\begin{lstlisting}[language=Java]
{
    "Staves": {
        "1": {
            "page_to_local": {
                // page-local COCO ID --> staff-local COCO ID
                // of the same physical object
                "123": "123",
                "456": "456",
                "789": "789",
                ...
            }
        },
        ...
    },
    "Grandstaves": {},
    "Systems": {}
}
\end{lstlisting}

Again, for each subdivision instance (\verb`"Staves"."1"`, for example), there is one JSON object (\verb`Staves/1/coco-object-detection` \verb`.json` in this case). This object has a single field \verb`page_to_local`:

\begin{verbatim}
{
    "page_to_local": {
        "123": "123",
        "456": "456",
        "789": "789",
        ...
    }
}
\end{verbatim}

The \verb`page_to_local` field is a mapping dictionary, that maps COCO annotation object IDs in the page-level \verb`coco-object-detection.json` file, to IDs in the subdivision-level \verb`coco-object-detection.json` file (staff, grandstaff, system).

\subsection{layout.json}

Each page may contain a \verb`layout.json` file. This is a COCO object-detection file with bounding box annotations of the following objects:

\begin{itemize}
    \item \verb`staff`: A staff that participates in music notation.
    \item \verb`emptyStaff`: An empty staff that does not contain any music, usually used for spacing, lyrics or song title. Empty staves inside systems (covered by brackets or braces) are not considered empty staves, instead are treated as regular staves with empty musical content. These are very common in particellas. This decision was made to aid harmonization with MuNG and MusicXML transcriptions.
    \item \verb`grandstaff`: A pair of staves that looks like a piano grandstaff (either has true pianoform music, or has the curly brace, or has the G-clef F-clef combination). It may also be a guitar, harp, or a pair of violins - appearance is more important then semantics.
    \item \verb`system`: A set of staves that play simultaneously (multiple instruments).
    \item \verb`staffMeasure`: One measure within a \verb`staff`. Empty space on a staff is NOT a staff measure.
    \item \verb`grandstaffMeasure`: One measure within a \verb`grandstaff`.
    \item \verb`systemMeasure` One measure within a \verb`system`.
\end{itemize}

\begin{figure}[p]
    \centering
    \includegraphics[width=0.9\linewidth]{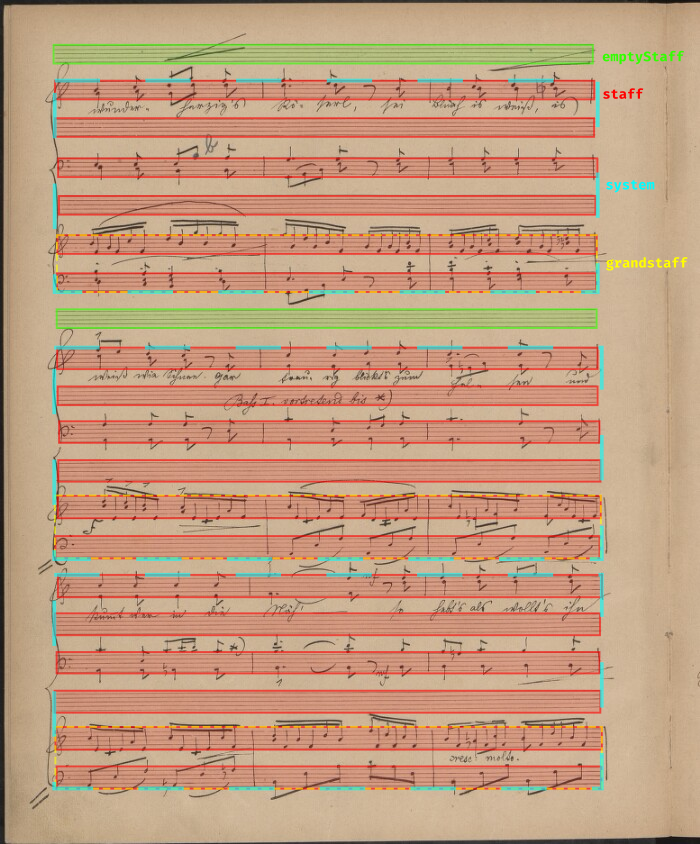}
    \caption{Staff-level annotations stored in the layout.json file.}
    \label{fig:layout-json}
\end{figure}

\begin{figure}[p]
    \centering
    \includegraphics[width=0.9\linewidth]{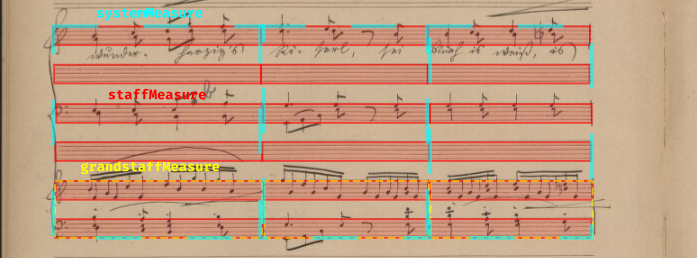}
    \caption{Measure-level annotations stored in the layout.json file.}
    \label{fig:layout-measures-json}
\end{figure}

The bounding boxes should be tightly wrapped around the annotated object (and fully contain it). An example annotated page with \verb`staff`, \verb`emptyStaff`, \verb`grandstaff` and \verb`system` annotations visualized can be seen in \autoref{fig:layout-json}. A single system of that page with \verb`staffMeasure`, \verb`grandstaffMeasure` and \verb`system` \verb`Measure` annotations can be seen in \autoref{fig:layout-measures-json}.

The contents of the \verb`layout.json` file are analogous to the \verb`coco-object-` \verb`detection.json` file, except for the annotations and its semantics. This is a shortened example for the \textit{Page} \verb`8136b106-6283-42c6-99eb-2f46c519c931_` \verb`d1769738-290b-4810-90b7-19fd8708d0c7` in the \verb`UFAL.OmniOMR` dataset:

\begin{lstlisting}[language=Java]
{
    "info": {
        "year": 2025,
        "version": "1.0",
        "description": "UFAL.OmniOMR",
        "contributor": "Institute of Formal and Applied Linguistics, Charles University",
        "url": "https://ufal.mff.cuni.cz/grants/omniomr",
        "date_created": "2025/11/28"
    },
    "licenses": [
        {
            "id": 0,
            "name": "UFAL.OmniOMR/LICENSE.txt",
            "url": "musicorpus://UFAL.OmniOMR/LICENSE.txt",
        }
    ],
    "images": [
        {
            "id": 0,
            "width": 3109,
            "height": 3751,
            "file_name": "8136b106-6283-42c6-99eb-2f46c519c931_d1769738-290b-4810-90b7-19fd8708d0c7/image.jpg",
            "license": 0,
            "date_captured": "2025-11-28 13:23:25"
        }
    ],
    "annotations": [
        {
            "id": 0,
            "image_id": 0,
            "category_id": 0,
            "segmentation": [[232, 347, 232, 451, 2637, 451, 2637, 347]],
            "area": 250120,
            "bbox": [232, 347, 2405, 104],
            "iscrowd": 0
        },
        ...
    ],
    "categories": [
        { "id": 0, "name": "staff" },
        { "id": 1, "name": "emptyStaff" },
        { "id": 2, "name": "grandstaff"},
        { "id": 3, "name": "system"},
        { "id": 4, "name": "staffMeasure"},
        { "id": 5, "name": "grandstaffMeasure"},
        { "id": 6, "name": "systemMeasure"}
    ]
}
\end{lstlisting}

\subsection{transcription.musicxml}

Each page, staff, grandstaff or system may contain a \verb`transcription.musicxml` file. MusicXML file is a weakly-aligned transcription for end-to-end recognition models. For more info about the format see \url{https://www.w3.org/2021/06/musicxml40/}. There are a few guidelines you should stick to when publishing MusicXML data in MusiCorpus:

\begin{itemize}[noitemsep]
    \item The files should not be compressed (no \verb`.mxl` files)
    \item It should be at least MusicXML 4.0.
    \item Line breaks should be encoded explicitly using the \verb`<print new-system=` \verb`"yes">` element.
    \item Invisible clefs and key/time signatures should be encoded explicitly with the \verb`print-object="no"` attribute.
\end{itemize}

\end{document}